\documentclass[10pt,twocolumn,letterpaper]{article}
\usepackage[pagenumbers]{cvpr} 
\definecolor{cvprblue}{rgb}{0.21,0.49,0.74}
\usepackage[pagebackref,breaklinks,colorlinks,allcolors=cvprblue]{hyperref}
\def\paperID{5786} 
\def\confName{CVPR}
\def\confYear{2025}

\usepackage{textcomp}
\usepackage{stfloats}
\usepackage[pagebackref,breaklinks,colorlinks]{hyperref}
\usepackage{url}
\usepackage{verbatim}
\usepackage[capitalize]{cleveref}
\usepackage{booktabs}
\usepackage{multirow}
\usepackage[normalem]{ulem}
\usepackage{float}
\usepackage{wrapfig}
\usepackage{amsmath}
\usepackage{amssymb}
\usepackage{bbding}
\usepackage{caption}
\usepackage{algorithm}
\usepackage{algpseudocode}
\usepackage{cleveref}
\usepackage{balance}
\usepackage{xcolor}
\useunder{\uline}{\ul}{}

\begin{document}
\title{AIM-Fair: Advancing Algorithmic Fairness via Selectively Fine-Tuning Biased Models with Contextual Synthetic Data}
\author{
Zengqun Zhao \hspace{1.5em} 
Ziquan Liu \hspace{1.5em} 
Yu Cao \hspace{1.5em}
Shaogang Gong \hspace{1.5em}
Ioannis Patras\\
Centre for Multimodal AI, Queen Mary University of London\\
{\tt\small \{zengqun.zhao, ziquan.liu, yu.cao, s.gong, i.patras\}@qmul.ac.uk}
}

\maketitle

\begin{abstract}
Recent advances in generative models have sparked research on improving model fairness with AI-generated data. However, existing methods often face limitations in the diversity and quality of synthetic data, leading to compromised fairness and overall model accuracy. Moreover, many approaches rely on the availability of demographic group labels, which are often costly to annotate. This paper proposes AIM-Fair, aiming to overcome these limitations and harness the potential of cutting-edge generative models in promoting algorithmic fairness. We investigate a fine-tuning paradigm starting from a biased model initially trained on real-world data without demographic annotations. This model is then fine-tuned using unbiased synthetic data generated by a state-of-the-art diffusion model to improve its fairness. Two key challenges are identified in this fine-tuning paradigm, 1) the low quality of synthetic data, which can still happen even with advanced generative models, and 2) the domain and bias gap between real and synthetic data. To address the limitation of synthetic data quality, we propose Contextual Synthetic Data Generation (CSDG) to generate data using a text-to-image diffusion model (T2I) with prompts generated by a context-aware LLM, ensuring both data diversity and control of bias in synthetic data. To resolve domain and bias shifts, we introduce a novel selective fine-tuning scheme in which only model parameters more sensitive to bias and less sensitive to domain shift are updated. Experiments on CelebA and UTKFace datasets show that our AIM-Fair improves model fairness while maintaining utility, outperforming both fully and partially fine-tuned approaches to model fairness. \textit{The code is available at \url{https://github.com/zengqunzhao/AIM-Fair}.} \footnote{Corresponding author: Ziquan Liu \{ziquan.liu@qmul.ac.uk\}.}
\end{abstract}

\begin{figure}[!t]
	\centering
	\includegraphics[scale=1.28]{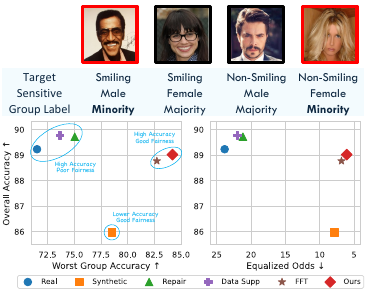}
    \vspace{-1em}
	\caption{Facial attributes classification (FAC) on the CelebA dataset based on different training strategies, in which \textbf{\textit{Smiling}} is the target attribute and \textbf{\textit{Gender}} is the protected attribute. This shows different learning strategies result in variable model utility (Overall Accuracy) vs. model fairness (Worst Group Accuracy and Equalized Odds) on demographic groups. A model trained solely on real data if biased exhibits high accuracy but poor fairness scores. Conversely, models trained on balanced synthetic data show better fairness but poorer accuracy due to a ``domain gap'' between the real and synthetic data, and a lack of selective model fine-tuning when synthetic data is deployed. Strategies to repair imbalances in real data \cite{ye2023exploiting,d2024improving} or to supplement real data with synthetic data \cite{shin2023fill} marginally increase accuracy but do little to improve fairness. Our method for selective fine-tuning of a pre-trained (biased) model with synthetic data not only preserves model accuracy but also substantially improves fairness, outperforming fully fine-tuning (FFT) in both model utility and fairness.}
	\label{fig_intro}
\vspace{-1.8em}
\end{figure}

\vspace{-0.5em}
\section{Introduction}
\vspace{-0.5em}
Recent research has raised significant concerns about fairness and bias in machine learning models \cite{mehrabi2021survey}. These models often demonstrate varying performance across different demographic groups, leading to unfair outcomes. To mitigate the spurious correlation caused by learning from imbalanced (biased) data, regularizations were used as optimization objectives \cite{arjovsky2019invariant,duchi2021learning,donini2018empirical,levy2020large}. Methods like distributionally robust optimization (DRO) \cite{duchi2021learning} optimizes the worst-case performance, while invariant risk minimization (IRM) \cite{arjovsky2019invariant} learns unbiased representations with invariance to different environments. Influenced by the success of representation learning \cite{uelwer2023survey}, attempts were made to learn a fair feature representation invariant to protected facial attributes \cite{creager2019flexibly,madras2018learning,park2021learning,wang2020towards}. However, these methods rely on demographic group labelling, often unavailable in practice \cite{ashurst2023fairness,choi2020fair}. More recent advances in generative models have utilized generative augmentation for a more balanced training data distribution \cite{zhang2024distributionally,d2024improving,dash2022evaluating,joo2020gender,ramaswamy2021fair,peychev2022latent}. These techniques primarily focus on image editing, either on real data \cite{ramaswamy2021fair,peychev2022latent} or synthetic data \cite{d2024improving}, to create less biased training data. Yet, these methods still require additional group annotations of the data they edit. To address this limitation, DiGA \cite{zhang2024distributionally} proposed to create an unbiased training set by editing spurious facial attributes to a random degree while keeping a target facial attribute unchanged without knowing each sample's group label. However, DiGA increases training data size multiple times, leading to substantially higher training costs, and the quality of generated data is limited due to the unknown group labels. 

\begin{figure*}[!t]
	\centering
        \vspace{-1.8em}
	\includegraphics[scale=0.51]{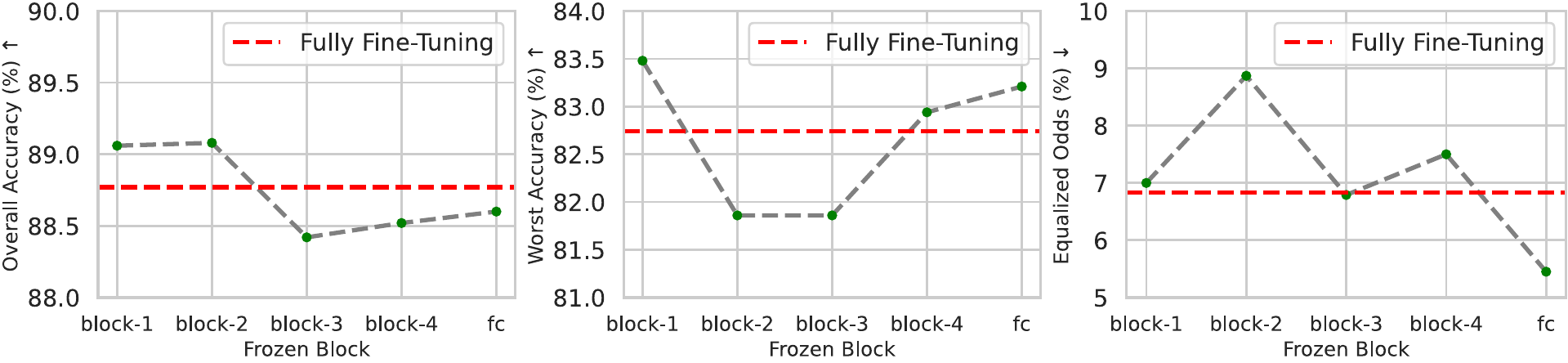}
        \vspace{-0.8em}
	\caption{The effects of selective fine-tuning by layer-wise freezing from facial attribute classifications on the CelebA dataset, with Smiling as the target attribute and Male as the protected attribute: When only the fully connected (FC) layer is frozen the model shows improved worst demographic group accuracy and reduced equalized odds, i.e. more fair, but sacrifices some utility (overall accuracy). This indicates increased cross-domain generalisability (real and synthetic). Conversely, freezing only block 2 while fine-tuning the remaining parameters results in high overall accuracy but poorer fairness, i.e. further enhanced domain-bias specificity. When only block 1 is frozen, the model not only maintained equalized odds but also increased utility (overall accuracy) and worst group accuracies.}
	\label{fig_layer_wise_results}
\vspace{-1.7em}
\end{figure*}

Recent advancements in text-to-image generative models showcase impressive data fidelity \cite{rombach2022high}, yet their potential for improving fairness through data expansion has not been fully explored by the fairness community. This raises the question: \textbf{can AI-generated synthetic data play a crucial role in mitigating biases within machine learning models?} This research question is particularly important given the successful application of AI-generated data in fine-tuning large language models (LLMs) \cite{dubey2024llama}, data augmentation \cite{zhou2023using}, and long-tail recognition \cite{shin2023fill}, alongside some reported limitations of synthetic data effectiveness \cite{shumailov2024ai}. This work presents a comprehensive empirical investigation into whether fine-tuning on high-quality, balanced generative data from a contemporary text-to-image model can counteract model biases caused by training on imbalanced real data. We identify two key challenges in bias-counter fine-tuning with synthetic data: (1) \textit{A data-related challenge arising from linguistic ambiguity of the textual prompt and/or model misrepresentation \cite{zhang2023iti,shrestha2024fairrag}, which results in low-quality and low-diversity generated data.} (2) \textit{A model learning challenge caused by both a domain shift (synthetic vs. real) and a bias shift (unbiased vs. biased) between the real and the synthetic data.} Fine-tuning blindly on the synthetic will result in a model with decreased utility.

To address the data quality issue, we propose Contextual Synthetic Data Generation (CSDG). Existing Latent Diffusion Models (LDMs) \cite{peychev2022latent,chenpixart,ramesh2022hierarchical,saharia2022} often use vision-language models, such as CLIP, for cross-modality alignment. To yield more diverse and better fine-grained image generation, we formulate a contextual synthetic data generation strategy that uses more detailed text descriptions from expansive linguistic expressions of LLMs to condition an LDM with richer, context-driven text. As a result, the generated images cover more scenarios and provide more details, enhancing diversity and mitigating bias in real data. In contrast to other methods that edit or manipulate real data samples, a key strength of our method is that it does not require annotating the Protected Group Attributes of real data.

The model learning challenge is due to the existence of two types of shifts between the generated and the real data, namely, the desired bias shift and the undesired domain shift (image quality, realism, scene characteristics). To address this challenge we propose to locate and update model parameters that are more sensitive to the bias shift and less sensitive to the domain shift. Our observation is that some parameters are more sensitive to data distribution shift (cross-domain generalisability) -- we call \textit{domain-sensitive parameters}, while others are more sensitive to demographic group discrepancies (real data domain bias specificity), which we call \textit{fairness-sensitive parameters}. This observation is supported by the experiments shown in Fig.~\ref{fig_layer_wise_results}. This is from fine-tuning a real data pre-trained model on balanced synthetic data while keeping parameters at different layers frozen. To identify which parameters to update, we propose a novel selection scheme in which the gradient differences between the updates by the real (biased) dataset and two synthetic datasets with one unbiased and another biased. Ranking the gradient differences between the synthetic-biased data and the synthetic-unbiased data reveals parameters sensitive to fairness (fairness-sensitive parameters). In contrast, inverse ranking the gradient differences between the real data and the synthetic-biased data discovers parameters less sensitive to domain shift (domain-insensitive parameters). In fine-tuning, a selection mask is constructed as the intersection of the top-k rankings. This selection mask is applied to initialize the pre-trained model, so that only the selected parameters are updated using the balanced synthetic data.

\begin{figure*}[!t]
    \centering
    \vspace{-2em}
    \includegraphics[scale=0.75]{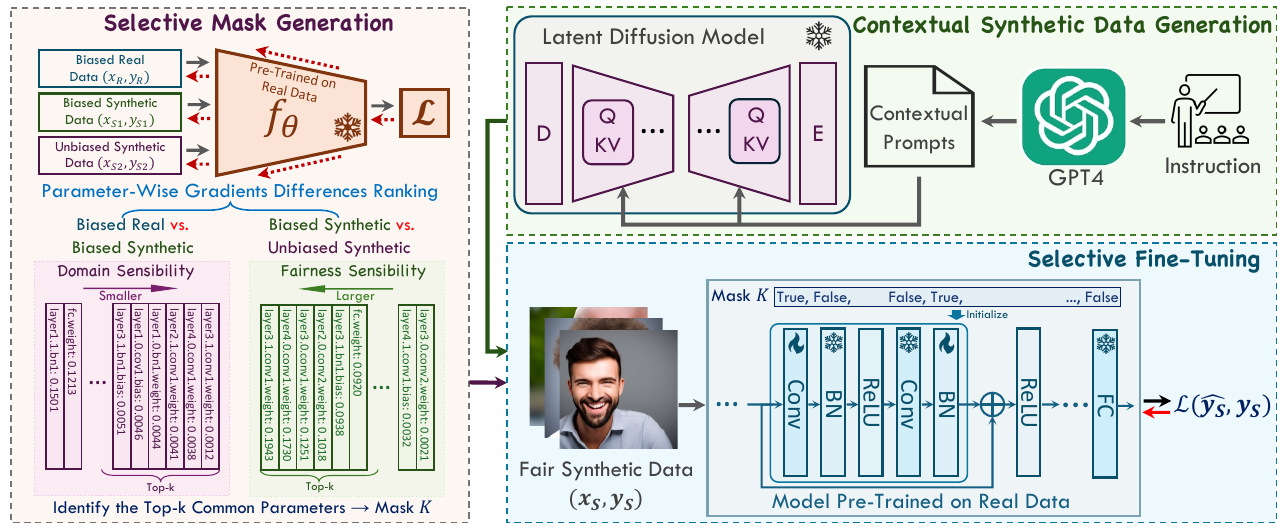}
    \vspace{-1em}
    \caption{A selective fine-tuning model consisting of three parts: (1) Contextual Synthetic Data Generation (CSDG) for generating diverse images using GPT-4 generated prompts, (2) Selective Mask Generation (SMG) for creating a selection mask that determines which parameters are updated during fine-tuning, and (3) Selective Fine-Tuning (SFT) to enhance model fairness obtained from synthetic data whilst simultaneously to preserve model utility yielded from real data in pre-training.}
    \label{fig_strcture}
    \vspace{-1.8em}
\end{figure*}

Fig.~\ref{fig_intro} presents a comparison of the results across various training strategies using real and/or synthetic data. Fig~\ref{fig_strcture} shows an overview of our model, which consists of three parts: Contextual Synthetic Data Generation (CSDG), Selective Mask Generation (SMG), and Selective Fine-Tuning (SFT). CSDG uses a pre-trained and fixed LDM to generate high-quality images, where a series of contextual prompts serve as the conditions. SMG creates a selection mask that determines which parameters are updated during fine-tuning. SFT is designed to correct bias in the model. We initialize a pre-trained model using the parameter selection mask. Then the model is fine-tuned on balanced synthetic data to enhance its fairness while retaining the model utility. Our contributions are as follows: 
\begin{itemize}
\item We investigate a fine-tuning paradigm that mitigates model bias stemming from unbalanced real data using synthetic data generated by a text-to-image (T2I) process without requiring demographic group annotations.
\item We design contextual synthetic data generating by using a T2I diffusion mode with prompts generated by a context-aware LLM, ensuring both data diversity and control of bias in synthetic data.
\item We introduce a selective fine-tuning method for fair model learning, which identifies domain-sensitive and fairness-sensitive parameters for improving model fairness and utility simultaneously. 
\item Our method outperforms both full and partial fine-tuning methods and achieves superior performance compared to state-of-the-art methods across several datasets.
\end{itemize}

\vspace{-0.5em}
\section{Related Work}
\vspace{-0.5em}

\noindent {\bf Mitigating Model Bias.}
Current methods for model fairness can be categorized into three types: \textit{pre-processing}, \textit{in-processing}, and \textit{post-processing}.
\textbf{\textit{Pre-processing}} approaches modify sample distributions of protected variables or perform transformations to remove discrimination from the data. Many recent works use generative models to create balanced, unbiased datasets \cite{peychev2022latent,ramaswamy2021fair,zhang2023fairness}.
\textbf{\textit{In-processing}} methods incorporate fairness metrics into model optimization to maximize both performance and fairness \cite{arjovsky2019invariant,duchi2021learning,donini2018empirical,levy2020large,kappiyath2025sebra,xue2024bmft}. Some studies explore mitigating bias without group annotations, such as Just Train Twice (JTT) \cite{liu2021just}, which upweights misclassified examples to improve worst-group performance, and Cross-Tisk Minimization (XRM) \cite{pezeshkidiscovering}, which trains twin classifiers to reduce spurious correlations.
\textbf{\textit{Post-processing}} methods apply transformations to model outputs to improve fairness, such as model calibration \cite{pleiss2017fairness,liu2019implicit,oldfield2025multilinear} and thresholding \cite{hardt2016equality,corbett2023measure} to align predicted positive outcomes with actual positive examples across groups.
Our method falls within the \textit{pre-processing} paradigm, but instead of simply balancing real data with synthetic data, we focus on using synthetic data to enhance model fairness.

\noindent {\bf Improving Model Fairness on Generative Data.} Generative models have advanced rapidly in recent years \cite{rombach2022high,saharia2022photorealistic}, with several studies exploring their use to improve model fairness through generative data \cite{joo2020gender,d2024improving,zhang2024distributionally}. Much of the prior research has focused on generating counterfactual samples to assess fairness \cite{dash2022evaluating,joo2020gender}, and generative methods have also made strides in bias mitigation by creating balanced, unbiased datasets \cite{peychev2022latent,ramaswamy2021fair,zhang2023fairness}. Instead of generating counterfactual samples based on real data, D’Incà \etal \cite{d2024improving} use a diffusion model for uncontrolled image generation, followed by manipulation of the synthetic images in a semantic space. However, these approaches require additional group annotations. Zhang \etal \cite{zhang2024distributionally} use generative models to identify spurious attributes that may bias the model, then edit each image to modify these attributes. They train a fair FAC model on the augmented dataset, enhancing its invariance to these spurious variations. While this approach does not require protected group labels for training the FAC model, annotations are still needed to train the generative models that detect and edit spurious attributes. Additionally, generating accurate counterfactual images remains challenging, as the edits are applied randomly to images without known protected attributes. In contrast, our method uses generative data from a text-to-image LDM, enabling greater control over the diversity of the generated data and allowing a broader range of scenarios and variations.

\noindent {\bf Model Fine-Tuning.}
A common approach to transfer learning in the presence of distribution shifts is to fine-tune the last few layers of a pre-trained model, retaining the learned features while adapting to the new task \cite{zhuang2020comprehensive,liu2023twins}. To prevent overfitting during this fine-tuning process, existing methods suggest using a smaller learning rate compared to the initial pretraining phase \cite{kornblith2019better}, freezing the early backbone layers and gradually unfreezing them \cite{howard2018universal}, or applying different learning rates to each layer \cite{ro2021autolr}. Lee et al. \cite{leesurgical} introduced surgical fine-tuning, which showed that selectively fine-tuning a subset of layers can either match or outperform conventional fine-tuning strategies. Recent research \cite{kaplun2023less} indicates that training a carefully selected subset of layers while keeping the remaining weights frozen at their initial values can lead to varying contributions from different layers across the network to overall performance. Our proposed method inherits some common findings from these works but we further investigated with parameters-wise selective fine-tuning, which details the parameters' sensibility to the property from the pre-train data distribution and the property from the downstream data distribution.

\vspace{-0.5em}
\section{Methods}
\vspace{-0.5em}
This section first introduces how to generate contextual synthetic data with LLM-generated prompts. Then, we detail how to obtain the parameters selection mask. Finally, we present how to conduct model fair fine-tuning with a selection mask on balanced synthetic data. We provide the algorithm for the selective fine-tuning in the \textit{Appendix}.

\vspace{-0.5em}
\subsection{Contextual Synthetic Data Generation}
\begin{figure}[!t]
	\centering
	\includegraphics[scale=0.85]{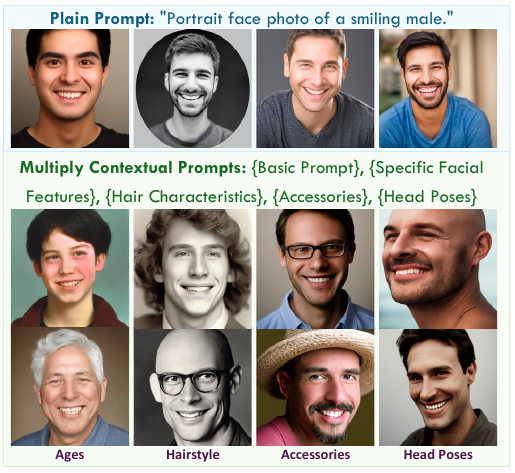}
        \vspace{-1em}
	\caption{Generated images of \textit{Smiling Male} conditioned on different prompts. Compared to the plain prompt, our contextual prompts enhance the diversity. More on UTKFace in \textit{Appendix}.} 
	\label{fig_vis}
\vspace{-2em}
\end{figure}

\vspace{-0.5em}
Our method attempts to correct the biased model trained on imbalanced real data by fine-tuning on balanced synthetic data, so the quality of the synthetic data is crucial to the success of fair learning. Although current text-to-image models have demonstrated remarkable performance, several studies indicate that directly expressing the desired attributes in the prompt often results in sub-optimal outcomes due to linguistic ambiguity or model misrepresentation \cite{zhang2023iti,shrestha2024fairrag}. For example, as shown in Fig.~\ref{fig_vis}, to generate a face photo with both target and protected attributes, one might use a prompt like “Portrait face photo of a smiling male.” as suggested in \cite{ye2023exploiting}. However, such manually designed prompts may lead to stereotypical image generation, excluding certain attributes or minority groups, which in turn can introduce bias in other attributes, such as age or hairstyle \cite{zhang2023iti}.

The recent work in zero-shot \cite{menonvisual} and supervised \cite{zhao2023dferclip} classification suggest that leveraging additional contextual and detailed information can enhance vision-language alignment. Considering current text-to-image models \cite{peychev2022latent,chenpixart,ramesh2022hierarchical,saharia2022} often use vision-language models, such as CLIP \cite{radford2021learning}, for cross-modality alignment, we propose a contextual synthetic data generation strategy that leverages the powerful linguistic capabilities of large language models, such as GPT-4, to condition an LDM with richer and context-driven text. As a result, the generated images cover more scenarios and provide more details, enhancing diversity and mitigating the bias in the characteristics not involved in the target and protected. The structure of the CSDG is shown in the upper right of Fig.~\ref{fig_strcture}. 

Concretely, the instruction provided for the GPT-4 model following the format: ``\{Task\}, \{Number of Prompts\}, \{Target Attribute\}, \{Protected Attribute\}, \{Other Detailed Descriptions\}, and \{Prompt Format\}''. For generating facial photos, for example, the other detailed descriptions contain specific facial features, hair characteristics, eye-related details, and head orientation or angle. Our text-to-image generation model is a pre-trained latent diffusion model--Stable Diffusion (SD) \cite{rombach2022high}, which reverses noises applied to the latent embedding of images. SD contains a variational autoencoder (VAE) \cite{lopez2018information}, a CLIP text encoder \cite{radford2021learning}, and a U-Net \cite{ronneberger2015u}. During the inference, the random Gaussian noise $\varepsilon_t \sim \mathcal{N}(0, \textbf{I})$ and the contextual prompt features $c = (w_1, w_2, \ldots, w_n)$ encoded via a CLIP text encoder $\Psi(\cdot)$ will be fed into the U-Net to condition the denoising process, where $n$ is the number of the contextual prompts. We provide detailed instruction and contextual prompts in the \textit{Appendix}. The empirical results in Tab.~\ref{tab_diff_prompts} demonstrate that the generated contextual synthetic data mitigate domain shift and improve model fairness.

\vspace{-0.5em}
\subsection{Selective Mask Generation}
\vspace{-0.5em}
As recent studies \cite{zhou2023using,wang2024domain,singh2024synthetic} have mentioned, synthetic data often fail to align with the real data distribution due to domain shift, suggesting that fine-tuning the pre-trained model directly on the balanced synthetic data may not learn the desired properties effectively. This is caused by the fact that there is both a domain shift (synthetic vs. real) and a bias shift (unbiased vs. biased) between the real and the synthetic data, and fine-tuning with the domain shift will result in a model with decreased utility. The empirical finding shown in Fig.~\ref{fig_layer_wise_results} demonstrates that when fine-tuning a real data pre-trained model on balanced synthetic data, some parameters are more sensitive to the data distribution shift, called \textit{domain-sensitive parameters}, while some are more sensitive to group discrepancies, called \textit{fairness-sensitive parameters}. To discover the parameters' sensibility towards different scenarios, we propose to construct three distinct datasets with different distributions to elicit the model responses by calculating the parameter-wise gradients.

We aim to fine-tune a pre-trained model \( f_\theta(x) \) to improve fairness while accounting for domain differences between real and synthetic data, hence, we want to find the parameters which are more sensitive to fairness while less sensitive to domain shift. Specifically, we start by constructing three different datasets: biased real data $ \{(x_i^{(R)}, y_i^{(R)})\}_{i=1}^{N_R} \in  \mathcal{D}_{R} $ representing training data with inherent biases, biased synthetic data $ \{(x_i^{(S_1)}, y_i^{(S_1)})\}_{i=1}^{N_{S_1}} \in  \mathcal{D}_{S_1} $ mirroring the unfairness of the real data, and unbiased synthetic data $ \{(x_i^{(S_2)}, y_i^{(S_2)})\}_{i=1}^{N_{S_2}} \in  \mathcal{D}_{S_2} $ designed to be fair, mitigating biases. $N_{R}$, $N_{S_1}$, and $N_{S_2} $ is the image number of the corresponding dataset.

We then compute the gradients $\textbf{\textit{g}}_R$, $\textbf{\textit{g}}_{S1}$, and $\textbf{\textit{g}}_{S2}$ of the loss function $\mathcal{L}(\theta; x, y)$, which is defined from a binary softmax loss. Rather than considering fine-grained, scalar-wise parameters, we select parameters at the level of weights and biases, denoted by $\theta$, which includes weights $W^{(l)}$ and biases $b^{(l)}$ for the Convolution Layer, Batch Normalization, and Fully Connected Layer, where $l$ refers to the layer index. Then, for the model parameters $\theta$ on each dataset:
\setlength{\belowdisplayskip}{1.6pt} \setlength{\belowdisplayshortskip}{1.6pt}
\setlength{\abovedisplayskip}{1.6pt} \setlength{\abovedisplayshortskip}{1.6pt}
\begin{equation}
\small
\vspace{-0.2em}
\begin{aligned}
\textbf{\textit{g}}_R   & = \frac{1}{N_R} \sum_{i=1}^{N_R} \nabla_{\boldsymbol{\theta}} \mathcal{L}(\theta; x_i^{(R)}, y_i^{(R)})
\end{aligned}
\vspace{-0.1em}
\end{equation}
\begin{equation}
\small
\vspace{-0.1em}
\begin{aligned}
\textbf{\textit{g}}_{S_1}   & = \frac{1}{N_{S_1}} \sum_{i=1}^{N_{S_1}} \nabla_{\theta} \mathcal{L}(\theta; x_i^{(S_1)}, y_i^{(S_1)})
\end{aligned}
\vspace{-0.1em}
\end{equation}
\begin{equation}
\small
\vspace{-0.1em}
\begin{aligned}
\textbf{\textit{g}}_{S_2}   & = \frac{1}{N_{S_2}} \sum_{i=1}^{N_{S_2}} \nabla_{\theta} \mathcal{L}(\theta; x_i^{(S_2)}, y_i^{(S_2)})
\end{aligned}
\vspace{-0.1em}
\end{equation}
where $\nabla$ is the gradient operator. We then calculate the gradient differences to capture how each parameter behaves across different data distributions. For parameter $\theta_j$:
\begin{equation}
\vspace{-0.1em}
\begin{aligned}
\Delta_{1,j} = \left| \textbf{\textit{g}}_{R,j} - \textbf{\textit{g}}_{S_1,j} \right| \\
\Delta_{2,j} = \left| \textbf{\textit{g}}_{S_1,j} - \textbf{\textit{g}}_{S_2,j} \right|
\end{aligned}
\vspace{-0.1em}
\end{equation}
where $\Delta_1$ measures parameters sensitive to the domain shift while $\Delta_2$ identifies parameters crucial for fairness, $j$ denotes the index of all the parameters. To obtain the parameters which are less affected by domain differences and more impactful for fairness, we conduct a ranking in ascending order on $\Delta_1$ (smaller differences first) and a ranking in descending order on $\Delta_2$ (larger differences first):
\begin{equation}
\vspace{-0.1em}
\begin{aligned}
R_1 = \text{argsort}(\Delta_1); R_2 = \text{argsort}(-\Delta_2)
\end{aligned}
\end{equation}

Then we can find the intersections from the \textit{top-k} parameters are both less sensitive to the domain gap and significant for fairness by $K = K_1 \cap K_2$, where
\begin{equation}
\vspace{-0.1em}
\begin{aligned}
K_1 = \{ \theta_j \mid j \in R_1[1:k] \} \\
K_2 = \{ \theta_j \mid j \in R_2[1:k] \}
\end{aligned}
\vspace{-0.1em}
\end{equation}

Finally, the selection mask can be obtained by:
\begin{equation}
\vspace{-0.3em}
\begin{aligned}
M_j = \begin{cases}
True, & \text{if } \theta_j \in K \\
False, & \text{otherwise}
\end{cases}
\end{aligned}
\vspace{-0.3em}
\end{equation}

\vspace{-0.3 em}
\subsection{Selective Fine-Tuning on Synthetic Data}
\vspace{-0.5 em}

Our goal is to fine-tune the pre-trained model $f_\theta(x)$ on the balanced synthetic dataset $\mathcal{D}_{S_2}$ following the ERM framework, while only updating selected parameters as indicated by the mask $M$. Specifically, given the model $f_\theta(x)$ pre-trained on the biased real data, balanced synthetic data $ \{(x_i^{(S)}, y_i^{(S)}\}_{i=1}^{N_{S_2}}$, and the parameters selection mask $M$, during optimization, we apply the mask to the gradients so that only the selected parameters are updated:
\begin{equation}
\begin{aligned}
\theta^{(t+1)} = \theta^{(t)} - \eta \left( M \odot \nabla_{\theta} \mathcal{L}\left( f_{\theta^{(t)}}(x_i^{(S_2)}), y_i^{(S_2)} \right) \right)
\label{eq:masked_update}
\end{aligned}
\end{equation}
where $\theta^{(t)}$ are the parameters at iteration $t$, $\eta$ is the learning rate, and $\odot$ denotes element-wise multiplication. Notably, once the mask $M$ is provided by SMG, it will be applied throughout the entire fine-tuning optimization process.

\begin{table*}[!t]
\vspace{-1em}
\begin{center}
\begin{minipage}[b]{0.48\textwidth}
\caption{Comparisons to other methods on the CelebA dataset under settings of varied target and protected attributes.}
\vspace{-1.0em}
\label{tab_comp_to_sota}
\setlength{\tabcolsep}{1pt}
\resizebox{\columnwidth}{!}{%
\begin{tabular}{@{}c|ccc|ccc|ccc@{}}
\toprule
\multirow{2}{*}{Methods} & \multicolumn{3}{c|}{T: Smiling,   P: Male} & \multicolumn{3}{c|}{T: Smiling,   P: Young} & \multicolumn{3}{c}{T: Young,   P: Male} \\ \cmidrule(l){2-10} 
 & ACC   (↑) & WST(↑) & EO   (↓) & ACC   (↑) & WST(↑) & EO   (↓) & ACC   (↑) & WST(↑) & EO   (↓) \\ \midrule
ERM \cite{he2016deep} & 88.20 & 70.10 & 25.30 & 88.30 & 71.50 & 15.60 & 77.70 & 42.00 & 52.00 \\
CVaR DRO \cite{levy2020large} & 87.30 & 74.00 & 22.80 & 87.00 & 76.10 & 13.90 & 75.40 & 42.30 & 48.80 \\
EIIL \cite{creager2021environment} & 87.90 & 75.60 & 19.70 & 87.90 & 72.50 & 13.30 & 77.50 & 45.60 & 39.20 \\
LfF \cite{nam2020learning} & 87.10 & 77.50 & 17.00 & 85.30 & 72.90 & 14.30 & 77.40 & 44.20 & 43.60 \\
JTT \cite{liu2021just} & 88.00 & 74.80 & 19.40 & 87.60 & 73.30 & 14.20 & 76.30 & 43.60 & 47.70 \\
MAPLE \cite{zhou2022model} & 88.10 & 72.00 & 19.60 & 88.10 & 73.60 & 13.60 & 76.30 & 46.20 & 43.50 \\
DiGA \cite{zhang2024distributionally} & {\ul 88.40} & {\ul 81.90} & {\ul 7.40} & {\ul 89.10} & {\ul 78.50} & {\ul 9.50} & \textbf{80.00} & {\ul 51.30} & {\ul 33.30} \\ \midrule
\textbf{AIM-Fair (Ours)} & \textbf{89.02} & \textbf{84.20} & \textbf{6.07} & \textbf{90.21} & \textbf{87.78} & \textbf{5.44} & {\ul 78.19} & \textbf{54.89} & \textbf{28.18} \\ \bottomrule
\end{tabular}
}
\end{minipage}
\hspace{0.5em}
\begin{minipage}[b]{0.48\textwidth}
\caption{Comparisons to other methods on the CelebA dataset (T=Smiling, P=Male) under settings of training set sizes.}
\vspace{-1.0em}
\label{tab_diff_real_data_number}
\setlength{\tabcolsep}{1pt}
\resizebox{\columnwidth}{!}{%
\begin{tabular}{@{}c|ccc|ccc|ccc@{}}
\toprule
\multirow{2}{*}{Methods} & \multicolumn{3}{c|}{Samples Ratio = 50\%}  & \multicolumn{3}{c|}{Samples Ratio = 25\%}  & \multicolumn{3}{c}{Samples Ratio = 10\%} \\ \cmidrule(l){2-10} 
         & ACC (↑) & WST(↑) & EO (↓) & ACC (↑) & WST(↑) & EO (↓) & ACC (↑)        & WST(↑) & EO (↓) \\ \midrule
ERM \cite{he2016deep}      & 87.50   & 67.80  & 26.10  & 87.10   & 65.90  & 27.70  & 86.90          & 62.80  & 28.90  \\
CVaR DRO \cite{levy2020large} & 86.60   & 72.90  & 22.10  & 86.60   & 72.30  & 22.40  & 85.50          & 69.10  & 27.30  \\
EIIL \cite{creager2021environment}     & 86.20   & 71.30  & 22.50  & 85.90   & 69.60  & 25.40  & 86.80          & 64.20  & 26.70  \\
LfF \cite{nam2020learning}      & 86.90   & 75.50  & 19.40  & 85.90   & 72.10  & 23.60  & 85.50          & 66.10  & 27.70  \\
JTT \cite{liu2021just}      & 87.30   & 72.90  & 20.10  & 86.70   & 71.10  & 20.60  & 86.80          & 67.10  & 23.10  \\
MAPLE \cite{zhou2022model}    & 87.40   & 73.70  & 23.80  & 87.00   & 72.70  & 24.20  & 85.60          & 69.20  & 27.10  \\
DiGA \cite{zhang2024distributionally}      & {\ul88.40}   & {\ul81.10}  & {\ul7.80}   & {\ul88.40}   & {\ul78.30}  & {\ul8.00}   & \textbf{88.30} & {\ul78.80}  & {\ul8.40}   \\ \midrule
\textbf{AIM-Fair (Ours)} & \textbf{88.85} & \textbf{83.99} & \textbf{6.28} & \textbf{88.89} & \textbf{82.54} & \textbf{7.59} & {\ul87.90} & \textbf{81.73} & \textbf{8.16} \\ \bottomrule
\end{tabular}
}
\end{minipage}
\end{center}
\vspace{-2.5em}
\end{table*}
\vspace{-0.5em}
\section{Experiments}
\vspace{-0.5em}
This section presents the experimental setup and results. We begin by describing the datasets, followed by the implementation details. The main results and ablation studies are presented at the end.

\vspace{-0.5em}
\subsection{Datasets}
\vspace{-0.5em}
\textbf{CelebA} \cite{liu2015faceattributes} contains over 200,000 facial images with 40 binary attribute annotations. Following the setting of the previous works \cite{park2022fair,zhang2024distributionally,zhang2023fairness}, we set Male and Young to protected attributes and selected Smiling and Young as the target attribute which has the highest correlation with the protected attributes. For each experiment, we randomly sample a biased subset as a training dataset with a size of 20,000 images, where the majority group and minority group have 90\% and 10\% of the sample size respectively. We report performance on the whole original test dataset. \textbf{UTK Face} \cite{zhifei2017cvpr} consists of over 20,000 facial images with three kinds of annotations: gender, age, and ethnicity. Following the experimental setup in the previous works \cite{joo2020gender,jung2022learning,zhang2024distributionally}, we define a binary spurious attribute ``Ethnicity'' based on whether the facial image is white or not. The task is to predict the Gender. We randomly sample a biased subset of 10,000 images, with the same bias degree as CelebA. We also construct a balanced and unbiased test dataset consisting of 3,200 images.

\vspace{-0.5em}
\subsection{Fairness Metrics}
\vspace{-0.5 em}
Our goal is to learn a fair and accurate model. For model utility, we present the \textit{overall accuracy (ACC)} and \textit{group accuracy}. For fairness, follow previous work \cite{zhang2023fairness,zhang2024distributionally,park2022fair} we use \textit{equalized odds (EO)} \cite{hardt2016equality}, defined as:
\begin{equation}
\begin{aligned}
\overline{\sum}_{\forall y, \hat{y}} \left| P_{s_0}(\hat{Y} = \hat{y} \mid Y = y) - P_{s_1}(\hat{Y} = \hat{y} \mid Y = y) \right|,
\end{aligned}
\end{equation}
where $\overline{\sum}$ is the averaged sum, $Y$ target label, $\hat{Y}$ classifier predictive label, and $s_0, s_1 \in S$ protected attributes. The \textit{worst-group accuracy (WST)} defined as:
\begin{equation}
\underset{\forall y \in \mathcal{Y}, \forall s \in S} {\min} P_s(\hat{Y} = y \mid Y = y)
\vspace{-0.3em}
\end{equation}
and \textit{group standard deviation (STD)} \cite{wang2020mitigating} defined as:
\begin{equation}
\underset{\forall y \in \mathcal{Y}, \forall s \in S} {std} P_s(\hat{Y} = y \mid Y = y)
\vspace{-0.3em}
\end{equation}

\vspace{-0.2 em}
\subsection{Implementation Details}
\vspace{-0.5 em}
Following previous work \cite{zhang2024distributionally}, we use ResNet-18 \cite{he2016deep} as the backbone for all experiments. We use stable diffusion v2.1 as our latent diffusion model for generating images. We generated 10,000 images for each group and then randomly sampled $N_{S}$. To be consistent with the real data image number $N_{R}$, we set the $N_{S}=N_{R}/G$, where $G$ is the group number. During the pre-training phase, we set the batch size to 128 and trained the model for 15 epochs. The initial learning rate was set to 0.01, which was then reduced by a factor of 0.01 at the 10th epoch. During the fine-tuning phase, we set the batch size to 128 and trained the model for 10 epochs. And the learning rate is searched from \{0.4, 0.5, 0.6\}. All models are optimized by the SGD optimizer and trained on the Tesla A100 GPU based on the open-source PyTorch platform. To obtain more stable and reliable results, we conducted all experiments 10 times with different random seeds and then used the average as the final result.

\begin{table*}[!t]
\vspace{-1.5em}
\begin{center}
\caption{Comparisons of varied training strategies on CelebA and UTKFace datasets.}
\vspace{-1.0em}
\label{tab_diff_train_stra}
\setlength{\tabcolsep}{2pt}
\resizebox{2\columnwidth}{!}{%
\begin{tabular}{@{}c|c|cccccccccccc|cccccc@{}}
\toprule
\multirow{3}{*}{Methods} & \multirow{3}{*}{Target} & \multicolumn{12}{c|}{CelebA} & \multicolumn{6}{c}{UTKFace} \\ \cmidrule(l){3-20} 
 &  & \multicolumn{2}{c|}{Protected} & \multicolumn{4}{c|}{T: Smiling; P: Male} & \multicolumn{2}{c|}{Protected} & \multicolumn{4}{c|}{T: Smiling; P: Young} & \multicolumn{2}{c|}{Protected} & \multicolumn{4}{c}{T: Female; P: White} \\ \cmidrule(l){3-20} 
 &  & P=0 & \multicolumn{1}{c|}{P=1} & ACC (↑) & WST (↑) & EO (↓) & \multicolumn{1}{c|}{STD (↓)} & P=0 & \multicolumn{1}{c|}{P=1} & ACC (↑) & WST (↑) & EO (↓) & STD (↓) & P=0 & \multicolumn{1}{c|}{P=1} & ACC (↑) & WST (↑) & EO (↓) & STD (↓) \\ \midrule
\multirow{2}{*}{Baseline \cite{he2016deep}} & T=0 & 83.57 & \multicolumn{1}{c|}{98.50} & \multirow{2}{*}{89.23} & \multirow{2}{*}{71.52} & \multirow{2}{*}{23.84} & \multicolumn{1}{c|}{\multirow{2}{*}{10.65}} & 78.72 & \multicolumn{1}{c|}{96.09} & \multirow{2}{*}{90.19} & \multirow{2}{*}{78.72} & \multirow{2}{*}{17.37} & \multirow{2}{*}{6.92} & 79.58 & \multicolumn{1}{c|}{96.16} & \multirow{2}{*}{88.86} & \multirow{2}{*}{79.58} & \multirow{2}{*}{16.59} & \multirow{2}{*}{7.26} \\
 & T=1 & 95.36 & \multicolumn{1}{c|}{71.52} &  &  &  & \multicolumn{1}{c|}{} & 94.37 & \multicolumn{1}{c|}{86.42} &  &  &  &  & 95.75 & \multicolumn{1}{c|}{83.96} &  &  &  &  \\ \midrule
\multirow{2}{*}{\begin{tabular}[c]{@{}c@{}}Trained on \\ Synthetic Data\end{tabular}} & T=0 & 86.82 & \multicolumn{1}{c|}{89.40} & \multirow{2}{*}{85.98} & \multirow{2}{*}{78.52} & \multirow{2}{*}{7.87} & \multicolumn{1}{c|}{\multirow{2}{*}{4.07}} & 85.45 & \multicolumn{1}{c|}{85.76} & \multirow{2}{*}{85.12} & \multirow{2}{*}{82.42} & \multirow{2}{*}{\textbf{3.50}} & \multirow{2}{*}{\textbf{1.34}} & 82.71 & \multicolumn{1}{c|}{88.12} & \multirow{2}{*}{85.44} & \multirow{2}{*}{82.71} & \multirow{2}{*}{5.50} & \multirow{2}{*}{\textbf{1.96}} \\
 & T=1 & 86.39 & \multicolumn{1}{c|}{78.52} &  &  &  & \multicolumn{1}{c|}{} & 82.42 & \multicolumn{1}{c|}{85.27} &  &  &  &  & 84.90 & \multicolumn{1}{c|}{86.04} &  &  &  &  \\ \midrule
\multirow{2}{*}{\begin{tabular}[c]{@{}c@{}}Data\\ Supplementation \cite{shin2023fill}\end{tabular}} & T=0 & 84.39 & \multicolumn{1}{c|}{98.39} & \multirow{2}{*}{\textbf{89.77}} & \multirow{2}{*}{73.67} & \multirow{2}{*}{21.96} & \multicolumn{1}{c|}{\multirow{2}{*}{9.76}} & 78.05 & \multicolumn{1}{c|}{95.36} & \multirow{2}{*}{{\ul 90.35}} & \multirow{2}{*}{78.05} & \multirow{2}{*}{17.31} & \multirow{2}{*}{7.02} & 82.12 & \multicolumn{1}{c|}{93.97} & \multirow{2}{*}{\textbf{90.25}} & \multirow{2}{*}{82.12} & \multirow{2}{*}{11.85} & \multirow{2}{*}{5.24} \\
 & T=1 & 95.40 & \multicolumn{1}{c|}{73.67} &  &  &  & \multicolumn{1}{c|}{} & 94.90 & \multicolumn{1}{c|}{87.62} &  &  &  &  & 95.62 & \multicolumn{1}{c|}{89.28} &  &  &  &  \\ \midrule
\multirow{2}{*}{\begin{tabular}[c]{@{}c@{}}Data\\ Repairing \cite{ye2023exploiting,d2024improving}\end{tabular}} & T=0 & 82.82 & \multicolumn{1}{c|}{98.10} & \multirow{2}{*}{{\ul 89.72}} & \multirow{2}{*}{75.05} & \multirow{2}{*}{21.18} & \multicolumn{1}{c|}{\multirow{2}{*}{9.50}} & 79.50 & \multicolumn{1}{c|}{95.19} & \multirow{2}{*}{\textbf{90.68}} & \multirow{2}{*}{79.50} & \multirow{2}{*}{15.69} & \multirow{2}{*}{6.30} & 81.15 & \multicolumn{1}{c|}{94.22} & \multirow{2}{*}{{\ul 89.42}} & \multirow{2}{*}{81.15} & \multirow{2}{*}{13.37} & \multirow{2}{*}{5.89} \\
 & T=1 & 96.04 & \multicolumn{1}{c|}{75.05} &  &  &  & \multicolumn{1}{c|}{} & 94.51 & \multicolumn{1}{c|}{88.35} &  &  &  &  & 95.71 & \multicolumn{1}{c|}{86.61} &  &  &  &  \\ \midrule
\multirow{2}{*}{Linear Prob} & T=0 & 84.55 & \multicolumn{1}{c|}{96.22} & \multirow{2}{*}{89.69} & \multirow{2}{*}{77.41} & \multirow{2}{*}{17.51} & \multicolumn{1}{c|}{\multirow{2}{*}{7.71}} & 85.20 & \multicolumn{1}{c|}{93.81} & \multirow{2}{*}{90.14} & \multirow{2}{*}{85.20} & \multirow{2}{*}{8.61} & \multirow{2}{*}{3.18} & 81.06 & \multicolumn{1}{c|}{91.51} & \multirow{2}{*}{88.40} & \multirow{2}{*}{81.06} & \multirow{2}{*}{10.59} & \multirow{2}{*}{5.14} \\
 & T=1 & 94.82 & \multicolumn{1}{c|}{77.41} &  &  &  & \multicolumn{1}{c|}{} & 90.43 & \multicolumn{1}{c|}{87.88} &  &  &  &  & 94.59 & \multicolumn{1}{c|}{86.42} &  &  &  &  \\ \midrule
\multirow{2}{*}{Fully Fine-Tuning} & T=0 & 88.81 & \multicolumn{1}{c|}{91.62} & \multirow{2}{*}{88.77} & \multirow{2}{*}{{\ul 82.74}} & \multirow{2}{*}{{\ul 6.83}} & \multicolumn{1}{c|}{\multirow{2}{*}{{\ul 3.31}}} & 85.98 & \multicolumn{1}{c|}{90.06} & \multirow{2}{*}{87.85} & \multirow{2}{*}{{\ul 85.98}} & \multirow{2}{*}{{\ul 4.37}} & \multirow{2}{*}{{\ul 1.68}} & 83.00 & \multicolumn{1}{c|}{88.04} & \multirow{2}{*}{87.90} & \multirow{2}{*}{{\ul 83.00}} & \multirow{2}{*}{{\ul 5.40}} & \multirow{2}{*}{2.99} \\
 & T=1 & 89.55 & \multicolumn{1}{c|}{82.74} &  &  &  & \multicolumn{1}{c|}{} & 86.02 & \multicolumn{1}{c|}{86.81} &  &  &  &  & 90.74 & \multicolumn{1}{c|}{89.83} &  &  &  &  \\ \midrule
\multirow{2}{*}{\begin{tabular}[c]{@{}c@{}}\textbf{AIM-Fair (Ours)}\end{tabular}} & T=0 & 88.16 & \multicolumn{1}{c|}{91.40} & \multirow{2}{*}{89.02} & \multirow{2}{*}{\textbf{84.20}} & \multirow{2}{*}{\textbf{6.07}} & \multicolumn{1}{c|}{\multirow{2}{*}{\textbf{2.74}}} & 88.19 & \multicolumn{1}{c|}{93.64} & \multirow{2}{*}{90.21} & \multirow{2}{*}{\textbf{87.78}} & \multirow{2}{*}{5.44} & \multirow{2}{*}{2.34} & 84.26 & \multicolumn{1}{c|}{89.08} & \multirow{2}{*}{88.30} & \multirow{2}{*}{\textbf{84.26}} & \multirow{2}{*}{\textbf{4.81}} & \multirow{2}{*}{{\ul 2.41}} \\
 & T=1 & 90.25 & \multicolumn{1}{c|}{84.20} &  &  &  & \multicolumn{1}{c|}{} & 88.98 & \multicolumn{1}{c|}{87.78} &  &  &  &  & 90.62 & \multicolumn{1}{c|}{89.25} &  &  &  &  \\ \bottomrule
\end{tabular}
}
\vspace{-2em}
\end{center}
\end{table*}
\begin{table*}[!t]
\begin{center}
\caption{Comparisons of varied partial fine-tuning strategies on CelebA and UTKFace datasets.}
\vspace{-1.0em}
\label{tab_diff_selection_stra}
\setlength{\tabcolsep}{2pt}
\resizebox{2\columnwidth}{!}{%
\begin{tabular}{@{}c|c|cccccccccccc|cccccc@{}}
\toprule
\multirow{3}{*}{Methods} & \multirow{3}{*}{Target} & \multicolumn{12}{c|}{CelebA} & \multicolumn{6}{c}{UTKFace} \\ \cmidrule(l){3-20} 
 &  & \multicolumn{2}{c|}{Protected} & \multicolumn{4}{c|}{T: Smiling; P: Male} & \multicolumn{2}{c|}{Protected} & \multicolumn{4}{c|}{T: Smiling; P: Young} & \multicolumn{2}{c|}{Protected} & \multicolumn{4}{c}{T: Female; P: White} \\ \cmidrule(l){3-20} 
 &  & P=0 & \multicolumn{1}{c|}{P=1} & ACC (↑) & WST (↑) & EO (↓) & \multicolumn{1}{c|}{STD (↓)} & P=0 & \multicolumn{1}{c|}{P=1} & ACC (↑) & WST (↑) & EO (↓) & STD (↓) & P=0 & \multicolumn{1}{c|}{P=1} & ACC (↑) & WST (↑) & EO (↓) & STD (↓) \\ \midrule
\multirow{2}{*}{\begin{tabular}[c]{@{}c@{}}Best Random\\ Selection\end{tabular}} & T=0 & 87.93 & \multicolumn{1}{c|}{92.81} & \multirow{2}{*}{{\ul 89.33}} & \multirow{2}{*}{82.17} & \multirow{2}{*}{9.14} & \multicolumn{1}{c|}{\multirow{2}{*}{4.09}} & 85.52 & \multicolumn{1}{c|}{91.53} & \multirow{2}{*}{88.95} & \multirow{2}{*}{85.52} & \multirow{2}{*}{6.02} & \multirow{2}{*}{{\ul 2.16}} & 83.50 & \multicolumn{1}{c|}{88.85} & \multirow{2}{*}{87.92} & \multirow{2}{*}{83.50} & \multirow{2}{*}{5.38} & \multirow{2}{*}{2.61} \\
 & T=1 & 91.31 & \multicolumn{1}{c|}{82.17} &  &  &  & \multicolumn{1}{c|}{} & 88.25 & \multicolumn{1}{c|}{87.63} &  &  &  &  & 90.30 & \multicolumn{1}{c|}{89.01} &  &  &  &  \\ \midrule
\multirow{2}{*}{\begin{tabular}[c]{@{}c@{}}Best Sub-Tuning \cite{kaplun2023less}\\ (By Updating One Block)\end{tabular}} & T=0 & 87.50 & \multicolumn{1}{c|}{94.75} & \multirow{2}{*}{\textbf{90.04}} & \multirow{2}{*}{80.64} & \multirow{2}{*}{12.45} & \multicolumn{1}{c|}{\multirow{2}{*}{5.52}} & 87.28 & \multicolumn{1}{c|}{94.63} & \multirow{2}{*}{\textbf{90.33}} & \multirow{2}{*}{87.28} & \multirow{2}{*}{7.34} & \multirow{2}{*}{3.00} & 83.05 & \multicolumn{1}{c|}{89.56} & \multirow{2}{*}{\textbf{89.01}} & \multirow{2}{*}{83.05} & \multirow{2}{*}{6.51} & \multirow{2}{*}{3.56} \\
 & T=1 & 93.09 & \multicolumn{1}{c|}{80.64} &  &  &  & \multicolumn{1}{c|}{} & 89.15 & \multicolumn{1}{c|}{87.32} &  &  &  &  & 92.07 & \multicolumn{1}{c|}{91.34} &  &  &  &  \\ \midrule
\multirow{2}{*}{\begin{tabular}[c]{@{}c@{}}Best Sub-Tuning \cite{kaplun2023less} \\ (By Freezing One Block)\end{tabular}} & T=0 & 88.69 & \multicolumn{1}{c|}{91.09} & \multirow{2}{*}{89.06} & \multirow{2}{*}{{\ul 83.48}} & \multirow{2}{*}{{\ul 7.00}} & \multicolumn{1}{c|}{\multirow{2}{*}{{\ul 3.00}}} & 86.55 & \multicolumn{1}{c|}{91.89} & \multirow{2}{*}{88.65} & \multirow{2}{*}{86.55} & \multirow{2}{*}{\textbf{5.34}} & \multirow{2}{*}{2.25} & 83.58 & \multicolumn{1}{c|}{88.34} & \multirow{2}{*}{87.42} & \multirow{2}{*}{83.58} & \multirow{2}{*}{\textbf{4.76}} & \multirow{2}{*}{\textbf{2.23}} \\
 & T=1 & 90.48 & \multicolumn{1}{c|}{83.48} &  &  &  & \multicolumn{1}{c|}{} & 87.08 & \multicolumn{1}{c|}{86.55} &  &  &  &  & 88.96 & \multicolumn{1}{c|}{88.81} &  &  &  &  \\ \midrule
\multirow{2}{*}{\begin{tabular}[c]{@{}c@{}}Selective Fine-Tuning\\ (Cosine Similarity)\end{tabular}} & T=0 & 86.07 & \multicolumn{1}{c|}{91.77} & \multirow{2}{*}{88.85} & \multirow{2}{*}{83.33} & \multirow{2}{*}{8.40} & \multicolumn{1}{c|}{\multirow{2}{*}{3.61}} & 87.43 & \multicolumn{1}{c|}{92.96} & \multirow{2}{*}{{\ul 90.26}} & \multirow{2}{*}{{\ul 87.43}} & \multirow{2}{*}{5.53} & \multirow{2}{*}{\textbf{2.05}} & 83.72 & \multicolumn{1}{c|}{88.82} & \multirow{2}{*}{88.21} & \multirow{2}{*}{{\ul 83.72}} & \multirow{2}{*}{5.10} & \multirow{2}{*}{2.64} \\
 & T=1 & 91.53 & \multicolumn{1}{c|}{83.33} &  &  &  & \multicolumn{1}{c|}{} & 89.52 & \multicolumn{1}{c|}{88.65} &  &  &  &  & 90.14 & \multicolumn{1}{c|}{90.15} &  &  &  &  \\ \midrule
\multirow{2}{*}{\begin{tabular}[c]{@{}c@{}}AIM-Fair (Ours)\\ (Absolute Difference)\end{tabular}} & T=0 & 88.16 & \multicolumn{1}{c|}{91.40} & \multirow{2}{*}{89.02} & \multirow{2}{*}{\textbf{84.20}} & \multirow{2}{*}{\textbf{6.07}} & \multicolumn{1}{c|}{\multirow{2}{*}{\textbf{2.74}}} & 88.19 & \multicolumn{1}{c|}{93.64} & \multirow{2}{*}{90.21} & \multirow{2}{*}{\textbf{87.78}} & \multirow{2}{*}{{\ul 5.44}} & \multirow{2}{*}{2.34} & 84.26 & \multicolumn{1}{c|}{89.08} & \multirow{2}{*}{{\ul 88.30}} & \multirow{2}{*}{\textbf{84.26}} & \multirow{2}{*}{{\ul 4.81}} & \multirow{2}{*}{{\ul 2.41}} \\
 & T=1 & 90.25 & \multicolumn{1}{c|}{84.20} &  &  &  & \multicolumn{1}{c|}{} & 88.98 & \multicolumn{1}{c|}{87.78} &  &  &  &  & 90.62 & \multicolumn{1}{c|}{89.25} &  &  &  &  \\ \bottomrule
\end{tabular}
}
\end{center}
\vspace{-2.5em}
\end{table*}

\vspace{-0.5em}
\subsection{Comparisons to State of the Art}
\vspace{-0.5em}

We compared our method against seven contemporary techniques, including the baseline ERM \cite{he2016deep} method and six debiasing models all of which do not require protected attribute labels: Two regularization-based methods (CVaR DRO \cite{levy2020large} and EIIL \cite{creager2021environment}); three reweighting-based methods (LfF \cite{nam2020learning}, JTT \cite{liu2021just}, and MAPLE \cite{zhou2022model}); and generative-model-based method (DiGA \cite{zhang2024distributionally}). Our comparison covers two settings: different target and protected attributes and varying numbers of target labels.  

Tab.\ref{tab_comp_to_sota} shows that ERM achieves good accuracy but suffers from significant unfairness. Other debiasing methods, while improving fairness to some extent, generally sacrifice accuracy. DiGA \cite{zhang2024distributionally} improves fairness while maintaining accuracy by using a generative model to edit real data and create more balanced training data. In contrast, our method generates images from scratch using a text-driven latent diffusion model. It is evident that our method outperforms the best of the existing models DiGA~ \cite{zhang2024distributionally} in both fairness and accuracy in all categories, except ``Young / Male'' where we come close second to DiGA. We outperform all other methods consistently. We also conducted comparisons with other methods under smaller training set sizes, with the subsampling ratio of  50\%, 25\%, and 10\%. Tab.~\ref{tab_diff_real_data_number} shows that our method outperforms all others consistently, except on ACC score for Lable Ratio at 10\% where we come close second to DiGA. Critically, our method maintains robust model utility and fairness even with varying amounts of real training data, with these improvements attributed to the selective updating with balanced synthetic data. 

\vspace{-0.5em}
\subsection{Ablation Analysis}
\vspace{-0.5em}
\label{sec:ablation}
\noindent\textbf{Comparisons of Different Training Strategies.}
To evaluate the effectiveness of the proposed selective fine-tuning, we first compare our method with several other strategies trained on different types of data. These include the baseline, which trains the model using conventional ERM on real data, training the model solely on synthetic data, supplementing real data with synthetic data \cite{shin2023fill}, and balancing the real data \cite{ye2023exploiting,d2024improving}. Furthermore, we compare common fine-tuning methods, such as linear probing and full fine-tuning. 

The results in Tab.~\ref{tab_diff_train_stra} indicate that both data supplementation and data repairing do not significantly improve fairness. Data supplementation, which combines biased real data with balanced synthetic data for co-training, makes it difficult for the model to learn fairness properties from the synthetic data because of the domain gap. While data repairing can create balanced training data, the domain shift between real and synthetic data limits the effectiveness of fair learning. Common transfer learning methods, such as linear probing and full fine-tuning, also face challenges due to domain shifts. Specifically, linear probing is affected by discrepancies in feature representations, while full fine-tuning improves fairness at the cost of reduced model utility. 

\noindent\textbf{Comparisons of Different Partial Fine-Tuning.}
We also compare our method with other partial fine-tuning approaches, including random selection and sub-tuning \cite{kaplun2023less}. For random selection, we set the updating ratios at 40\%, 55\%, 70\%, and 85\%, and select the best result as the final one. For sub-tuning, we perform block-wise fine-tuning and block-wise freezing (i.e., updating one block or freezing one block while updating the rest), also selecting the block with the best result. 

Tab.~\ref{tab_diff_selection_stra} shows that fine-tuning only one block consistently yields the best accuracy, but at the cost of fairness, while freezing one block tends to improve fairness but often sacrifices model utility. This demonstrates that coarse-grained approaches, such as block-wise updating or freezing, struggle to achieve an optimal balance between model utility and fairness. In contrast, our method performs fine-grained, parameter-wise updating, which allows the model to enhance fairness while maintaining utility. As a result, our method achieves the best worst-group accuracy while preserving high overall accuracy.

Additionally, our method uses gradient differences to identify parameter sensitivity to varying data distributions. We also compared absolute gradient differences with the cosine similarity between gradients. The results in Tab.~\ref{tab_diff_selection_stra} show that using absolute gradient differences yields better results. The possible reason is that the cosine similarity only provides the direction of gradient disparity, whereas the absolute difference directly captures the magnitude of the gradient disparity.

\begin{table}[!t]
\begin{center}
\caption{Classification results on CelebA dataset (T=Smiling, P=Male) under settings of different prompt types and numbers.}
\vspace{-1.0em}
\label{tab_diff_prompts}
\setlength{\tabcolsep}{2.5pt}
\resizebox{\columnwidth}{!}{%
\begin{tabular}{c|c|cc|cccc|cc|cccc}
\toprule
\multirow{2}{*}{\begin{tabular}[c]{@{}c@{}}Prompt Types\\ (Number)\end{tabular}} & \multirow{2}{*}{Targe} & \multicolumn{2}{c|}{Protected} & \multicolumn{4}{c|}{T: Smiling; P: Male} & \multicolumn{2}{c|}{Protected} & \multicolumn{4}{c}{T: Smiling; P: Young} \\ \cmidrule{3-14} 
 &  & P=0 & P=1 & \begin{tabular}[c]{@{}c@{}}ACC\\ (↑)\end{tabular} & \begin{tabular}[c]{@{}c@{}}WST\\ (↑)\end{tabular} & \begin{tabular}[c]{@{}c@{}}EO\\ (↓)\end{tabular} & \begin{tabular}[c]{@{}c@{}}STD\\ (↓)\end{tabular} & P=0 & P=1 & \begin{tabular}[c]{@{}c@{}}ACC\\ (↑)\end{tabular} & \begin{tabular}[c]{@{}c@{}}WST\\ (↑)\end{tabular} & \begin{tabular}[c]{@{}c@{}}EO\\ (↓)\end{tabular} & \begin{tabular}[c]{@{}c@{}}STD\\ (↓)\end{tabular} \\ \midrule
\multirow{2}{*}{\begin{tabular}[c]{@{}c@{}}Plain\\ Prompt (1)\end{tabular}} & t=0 & 65.18 & 89.44 & \multirow{2}{*}{71.65} & \multirow{2}{*}{43.28} & \multirow{2}{*}{34.20} & \multirow{2}{*}{17.07} & 83.10 & 77.72 & \multirow{2}{*}{72.75} & \multirow{2}{*}{59.53} & \multirow{2}{*}{9.30} & \multirow{2}{*}{8.96} \\
 & t=1 & 77.47 & 43.28 &  &  &  &  & 59.53 & 68.83 &  &  &  &  \\ \midrule
\multirow{2}{*}{\begin{tabular}[c]{@{}c@{}}Contextual\\ Prompts (25)\end{tabular}} & t=0 & 77.12 & 89.25 & \multirow{2}{*}{81.85} & \multirow{2}{*}{69.73} & \multirow{2}{*}{16.3} & \multirow{2}{*}{7.65} & 76.72 & 67.90 & \multirow{2}{*}{79.80} & \multirow{2}{*}{67.90} & \multirow{2}{*}{8.82} & \multirow{2}{*}{8.75} \\
 & t=1 & 85.99 & 69.73 &  &  &  &  & 85.16 & 91.09 &  &  &  &  \\ \midrule
\multirow{2}{*}{\begin{tabular}[c]{@{}c@{}}Contextual\\ Prompts (50)\end{tabular}} & t=0 & 89.47 & 89.80 & \multirow{2}{*}{82.66} & \multirow{2}{*}{72.22} & \multirow{2}{*}{\textbf{5.07}} & \multirow{2}{*}{7.66} & 87.72 & 90.43 & \multirow{2}{*}{\textbf{85.85}} & \multirow{2}{*}{79.67} & \multirow{2}{*}{4.36} & \multirow{2}{*}{4.20} \\
 & t=1 & 77.26 & 72.22 &  &  &  &  & 79.67 & 82.67 &  &  &  &  \\ \midrule
\multirow{2}{*}{\begin{tabular}[c]{@{}c@{}}Contextual + \\ Head Poses (50)\end{tabular}} & t=0 & 86.82 & 89.40 & \multirow{2}{*}{\textbf{85.98}} & \multirow{2}{*}{\textbf{78.52}} & \multirow{2}{*}{7.87} & \multirow{2}{*}{\textbf{4.07}} & 85.45 & 85.76 & \multirow{2}{*}{85.12} & \multirow{2}{*}{\textbf{82.42}} & \multirow{2}{*}{\textbf{3.50}} & \multirow{2}{*}{\textbf{1.34}} \\
 & t=1 & 86.39 & 78.52 &  &  &  &  & 82.42 & 85.27 &  &  &  &  \\ \bottomrule
\end{tabular}
}
\end{center}
\vspace{-3em}
\end{table}

\begin{figure}[!t]
	\centering
        \vspace{-1em}
	\includegraphics[scale=0.56]{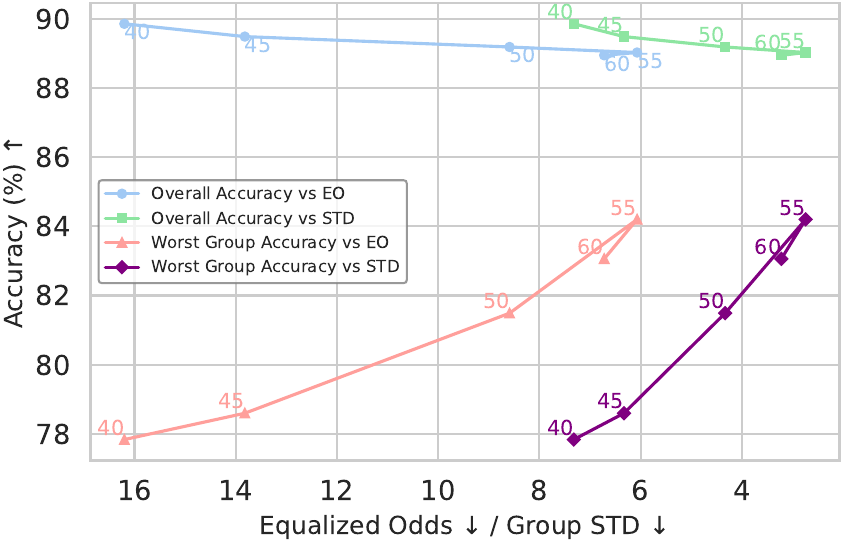}
        \vspace{-0.5em}
	\caption{Classification results on the CelebA dataset (T=Smiling, P=Male) with different top-k values. The top-k values \{40, 45, 50, 55, 60\} correspond approximately to \{64\%, 72\%, 80\%, 88\%, 96\%\} of the total model parameters.}
	\label{fig_diff_k}
\vspace{-2em}
\end{figure}

\noindent\textbf{Comparisons of Different Prompts.}
To evaluate the performance of the varied prompts used for generating images, we trained the model solely on the balanced synthetic data generated with corresponding prompts and tested it on the real test set. Tab.~\ref{tab_diff_prompts}  shows the results of different prompt types and quantities, indicating that images generated with contextual prompts lead to significantly better performance in both model accuracy and fairness. We believe this improvement is due to the increased diversity and details in the synthetic images generated from contextual prompts. Moreover, as the number of contextual prompts increases, test performance improves as well. Specifically, incorporating head pose variation into the prompt further enhances both accuracy and fairness in the generated images, as the results shown in the last row of Tab.~\ref{tab_diff_prompts}.

\noindent\textbf{Evaluations of Varied Top-k Values.}
In our method, to identify the intersection parameters from the two gradient difference rankings, we use the top-k selection to ensure that the chosen parameters are more sensitive to fairness and less affected by domain shift. The results for different top-k values are shown in Fig.~\ref{fig_diff_k}. When k is set to 55, our method achieves the best fairness performance while retaining overall accuracy. Intuitively, a lower k results in fewer updates, leading to better model utility but worse fairness, while a high k involves more parameter updates, which can negatively impact both accuracy and fairness.

\noindent\textbf{Evaluations of Varied Bias Ratio for Synthetic Data Construction.}
To evaluate the model’s sensitivity to domain shift and group disparity, we construct three distinct datasets with different distributions. In our method, the bias ratio for the biased synthetic data distribution is treated as a hyperparameter, referenced by the error set of real training examples, as suggested in JTT \cite{liu2021just}. To assess the impact of varying bias ratios in synthetic data, we explore different ratios. As shown in Tab.~\ref{tab_diff_syn_bias_ratio}, a disparity between the synthetic data bias ratio and the real data bias ratio leads to worse performance in both accuracy and fairness. However, despite these differences in bias ratios, our method still outperforms the fully fine-tuning approach. This demonstrates that our method is able to identify the model’s sensitivity under the bias ratios different to real data. 

\begin{table}[!t]
\vspace{-1em}
\begin{center}
\caption{Classification results on CelebA dataset (T=Smiling, P=Male) with varied bias ratio of the biased synthetic data.}
\vspace{-1.0em}
\label{tab_diff_syn_bias_ratio}
\setlength{\tabcolsep}{6pt}
\resizebox{0.9\columnwidth}{!}{%
\begin{tabular}{@{}c|c|cc|cccc@{}}
\toprule
\multirow{2}{*}{\begin{tabular}[c]{@{}c@{}}Bias\\ Ratio\end{tabular}} & \multirow{2}{*}{Target} & \multicolumn{2}{c|}{Protected} & \multirow{2}{*}{\begin{tabular}[c]{@{}c@{}}ACC\\ (↑)\end{tabular}} & \multirow{2}{*}{\begin{tabular}[c]{@{}c@{}}WST\\ (↑)\end{tabular}} & \multirow{2}{*}{\begin{tabular}[c]{@{}c@{}}EOD\\ (↓)\end{tabular}} & \multirow{2}{*}{\begin{tabular}[c]{@{}c@{}}STD\\ (↓)\end{tabular}} \\ \cmidrule(lr){3-4}
 &  & P=0 & P=1 &  &  &  &  \\ \midrule
\multirow{2}{*}{\begin{tabular}[c]{@{}c@{}}Fully\\ Fine-Tuning\end{tabular}} & t=0 & 88.81 & 91.62 & \multirow{2}{*}{88.77} & \multirow{2}{*}{82.74} & \multirow{2}{*}{6.83} & \multirow{2}{*}{3.31} \\
 & t=1 & 89.55 & 82.74 &  &  &  &  \\ \midrule
\multirow{2}{*}{4 : 6} & t=0 & 85.72 & 91.71 & \multirow{2}{*}{\textbf{89.23}} & \multirow{2}{*}{83.90} & \multirow{2}{*}{8.78} & \multirow{2}{*}{3.77} \\
 & t=1 & 92.69 & 83.90 &  &  &  &  \\ \midrule
\multirow{2}{*}{3 : 7} & t=0 & 87.05 & 91.62 & \multirow{2}{*}{88.90} & \multirow{2}{*}{83.52} & \multirow{2}{*}{7.72} & \multirow{2}{*}{3.26} \\
 & t=1 & 90.93 & 83.52 &  &  &  &  \\ \midrule
\multirow{2}{*}{2 : 8} & t=0 & 88.01 & 91.76 & \multirow{2}{*}{88.97} & \multirow{2}{*}{83.16} & \multirow{2}{*}{7.30} & \multirow{2}{*}{3.28} \\
 & t=1 & 90.46 & 83.16 &  &  &  &  \\ \midrule
\multirow{2}{*}{1 : 9} & t=0 & 88.16 & 91.40 & \multirow{2}{*}{89.02} & \multirow{2}{*}{\textbf{84.20}} & \multirow{2}{*}{\textbf{6.07}} & \multirow{2}{*}{\textbf{2.74}} \\
 & t=1 & 90.25 & 84.20 &  &  &  &  \\ \bottomrule
\end{tabular}
}
\end{center}
\vspace{-1.5em}
\end{table}
\begin{table}[!t]
\begin{center}
\caption{Classification results on CelebA dataset (T=Smiling, P=Male) with different number of synthetic data.}
\vspace{-1em}
\label{tab_diff_syn_data_number}
\setlength{\tabcolsep}{6pt}
\resizebox{0.9\columnwidth}{!}{%
\begin{tabular}{@{}c|c|cc|cccc@{}}
\toprule
\multirow{2}{*}{\begin{tabular}[c]{@{}c@{}}Ratio To\\ Real Data\end{tabular}} & \multirow{2}{*}{Target} & \multicolumn{2}{c|}{Protected} & \multirow{2}{*}{\begin{tabular}[c]{@{}c@{}}ACC\\ (↑)\end{tabular}} & \multirow{2}{*}{\begin{tabular}[c]{@{}c@{}}WST\\ (↑)\end{tabular}} & \multirow{2}{*}{\begin{tabular}[c]{@{}c@{}}EOD\\ (↓)\end{tabular}} & \multirow{2}{*}{\begin{tabular}[c]{@{}c@{}}STD\\ (↓)\end{tabular}} \\ \cmidrule(lr){3-4}
 &  & P=0 & P=1 &  &  &  &  \\ \midrule
\multirow{2}{*}{0.5} & t=0 & 88.65 & 93.96 & \multirow{2}{*}{\textbf{89.51}} & \multirow{2}{*}{80.80} & \multirow{2}{*}{10.31} & \multirow{2}{*}{4.90} \\
 & t=1 & 91.10 & 80.80 &  &  &  &  \\ \midrule
\multirow{2}{*}{1} & t=0 & 88.16 & 91.40 & \multirow{2}{*}{89.02} & \multirow{2}{*}{\textbf{84.20}} & \multirow{2}{*}{\textbf{6.07}} & \multirow{2}{*}{\textbf{2.74}} \\
 & t=1 & 90.25 & 84.20 &  &  &  &  \\ \midrule
\multirow{2}{*}{1.5} & t=0 & 89.18 & 92.05 & \multirow{2}{*}{88.34} & \multirow{2}{*}{81.25} & \multirow{2}{*}{7.13} & \multirow{2}{*}{3.98} \\
 & t=1 & 88.38 & 81.25 &  &  &  &  \\ \midrule
\multirow{2}{*}{2} & t=0 & 89.01 & 91.26 & \multirow{2}{*}{88.81} & \multirow{2}{*}{83.36} & \multirow{2}{*}{6.08} & \multirow{2}{*}{2.96} \\
 & t=1 & 91.26 & 89.44 &  &  &  &  \\ \bottomrule
\end{tabular}
}
\end{center}
\vspace{-2.5em}
\end{table}

\noindent\textbf{Evaluations of Different Number of Synthetic Data.} We also evaluate fine-tuning with varying amounts of balanced synthetic data. As shown in Tab.~\ref{tab_diff_syn_data_number}, we use the real training data count as a reference and set different ratios to determine the number of synthetic data. The results indicate that when the amount of synthetic data matches that of real data, the model achieves the best fairness. Additionally, using half the amount of real data results in the best accuracy. We believe that using too much synthetic data during fine-tuning can lead to overfitting, while using too little data may fail to adequately debias the model.

\vspace{-0.5 em}
\section{Conclusion}
\vspace{-0.5 em}
In this work, we proposed a method to mitigate bias in machine learning models using synthetic data generated by a text-to-image process. By designing contextual synthetic data generation and selective fine-tuning, we enhance model fairness without requiring demographic group annotations. Our model updates selectively fairness-sensitive parameters, optimizing simultaneously model fairness and utility scores. Empirical results demonstrate that our method outperforms existing techniques, improving fairness while maintaining model utility performance. This work highlights the potential of synthetic data for creating fairer AI systems, offering a promising direction for future research in bias mitigation.

\noindent \textbf{\textit{Acknowledgments}}: This research utilised Queen Mary's Apocrita HPC facility, supported by QMUL Research-IT. Zengqun Zhao is funded by Queen Mary Principal's PhD Studentships. Zengqun Zhao wants to thank Yining Wang and James Oldfield for the valuable comments and help, and Yiming Lin and Jie Shen for the valuable discussions.

{\small
\bibliographystyle{ieeenat_fullname}
\bibliography{reference}}

\begin{thebibliography}{65}
\providecommand{\natexlab}[1]{#1}
\providecommand{\url}[1]{\texttt{#1}}
\expandafter\ifx\csname urlstyle\endcsname\relax
  \providecommand{\doi}[1]{doi: #1}\else
  \providecommand{\doi}{doi: \begingroup \urlstyle{rm}\Url}\fi

\bibitem[Arjovsky et~al.(2019)Arjovsky, Bottou, Gulrajani, and Lopez-Paz]{arjovsky2019invariant}
Martin Arjovsky, L{\'e}on Bottou, Ishaan Gulrajani, and David Lopez-Paz.
\newblock Invariant risk minimization.
\newblock \emph{arXiv preprint arXiv:1907.02893}, 2019.

\bibitem[Ashurst and Weller(2023)]{ashurst2023fairness}
Carolyn Ashurst and Adrian Weller.
\newblock Fairness without demographic data: A survey of approaches.
\newblock In \emph{EAAMO}, pages 1--12, 2023.

\bibitem[Chen et~al.(2023)Chen, Jincheng, Chongjian, Yao, Xie, Wang, Kwok, Luo, Lu, and Li]{chenpixart}
Junsong Chen, YU Jincheng, GE Chongjian, Lewei Yao, Enze Xie, Zhongdao Wang, James Kwok, Ping Luo, Huchuan Lu, and Zhenguo Li.
\newblock Pixart-$\alpha$: Fast training of diffusion transformer for photorealistic text-to-image synthesis.
\newblock In \emph{ICLR}, 2023.

\bibitem[Choi et~al.(2020)Choi, Grover, Singh, Shu, and Ermon]{choi2020fair}
Kristy Choi, Aditya Grover, Trisha Singh, Rui Shu, and Stefano Ermon.
\newblock Fair generative modeling via weak supervision.
\newblock In \emph{ICML}, pages 1887--1898, 2020.

\bibitem[Corbett-Davies et~al.(2023)Corbett-Davies, Gaebler, Nilforoshan, Shroff, and Goel]{corbett2023measure}
Sam Corbett-Davies, Johann~D Gaebler, Hamed Nilforoshan, Ravi Shroff, and Sharad Goel.
\newblock The measure and mismeasure of fairness.
\newblock \emph{JMLR}, 24\penalty0 (1):\penalty0 14730--14846, 2023.

\bibitem[Creager et~al.(2019)Creager, Madras, Jacobsen, Weis, Swersky, Pitassi, and Zemel]{creager2019flexibly}
Elliot Creager, David Madras, J{\"o}rn-Henrik Jacobsen, Marissa Weis, Kevin Swersky, Toniann Pitassi, and Richard Zemel.
\newblock Flexibly fair representation learning by disentanglement.
\newblock In \emph{ICML}, pages 1436--1445, 2019.

\bibitem[Creager et~al.(2021)Creager, Jacobsen, and Zemel]{creager2021environment}
Elliot Creager, J{\"o}rn-Henrik Jacobsen, and Richard Zemel.
\newblock Environment inference for invariant learning.
\newblock In \emph{ICML}, pages 2189--2200, 2021.

\bibitem[Dash et~al.(2022)Dash, Balasubramanian, and Sharma]{dash2022evaluating}
Saloni Dash, Vineeth~N Balasubramanian, and Amit Sharma.
\newblock Evaluating and mitigating bias in image classifiers: A causal perspective using counterfactuals.
\newblock In \emph{WACV}, pages 915--924, 2022.

\bibitem[D'Inc{\`a} et~al.(2024)D'Inc{\`a}, Tzelepis, Patras, and Sebe]{d2024improving}
Moreno D'Inc{\`a}, Christos Tzelepis, Ioannis Patras, and Nicu Sebe.
\newblock Improving fairness using vision-language driven image augmentation.
\newblock In \emph{WACV}, pages 4695--4704, 2024.

\bibitem[Donini et~al.(2018)Donini, Oneto, Ben-David, Shawe-Taylor, and Pontil]{donini2018empirical}
Michele Donini, Luca Oneto, Shai Ben-David, John~S Shawe-Taylor, and Massimiliano Pontil.
\newblock Empirical risk minimization under fairness constraints.
\newblock \emph{NeurIPS}, 31, 2018.

\bibitem[Dubey et~al.(2024)Dubey, Jauhri, Pandey, Kadian, Al-Dahle, Letman, Mathur, Schelten, Yang, Fan, et~al.]{dubey2024llama}
Abhimanyu Dubey, Abhinav Jauhri, Abhinav Pandey, Abhishek Kadian, Ahmad Al-Dahle, Aiesha Letman, Akhil Mathur, Alan Schelten, Amy Yang, Angela Fan, et~al.
\newblock The llama 3 herd of models.
\newblock \emph{arXiv preprint arXiv:2407.21783}, 2024.

\bibitem[Duchi and Namkoong(2021)]{duchi2021learning}
John~C Duchi and Hongseok Namkoong.
\newblock Learning models with uniform performance via distributionally robust optimization.
\newblock \emph{The Annals of Statistics}, 49\penalty0 (3):\penalty0 1378--1406, 2021.

\bibitem[Hardt et~al.(2016)Hardt, Price, and Srebro]{hardt2016equality}
Moritz Hardt, Eric Price, and Nati Srebro.
\newblock Equality of opportunity in supervised learning.
\newblock \emph{NeurIPS}, 29, 2016.

\bibitem[He et~al.(2016)He, Zhang, Ren, and Sun]{he2016deep}
Kaiming He, Xiangyu Zhang, Shaoqing Ren, and Jian Sun.
\newblock Deep residual learning for image recognition.
\newblock In \emph{CVPR}, pages 770--778, 2016.

\bibitem[Howard and Ruder(2018)]{howard2018universal}
Jeremy Howard and Sebastian Ruder.
\newblock Universal language model fine-tuning for text classification.
\newblock In \emph{ACL}, pages 328--339, 2018.

\bibitem[Hu et~al.(2022)Hu, Shen, Wallis, Allen-Zhu, Li, Wang, Wang, Chen, et~al.]{hu2022lora}
Edward~J Hu, Yelong Shen, Phillip Wallis, Zeyuan Allen-Zhu, Yuanzhi Li, Shean Wang, Lu Wang, Weizhu Chen, et~al.
\newblock Lora: Low-rank adaptation of large language models.
\newblock In \emph{ICLR}, pages 1--13, 2022.

\bibitem[Joo and K{\"a}rkk{\"a}inen(2020)]{joo2020gender}
Jungseock Joo and Kimmo K{\"a}rkk{\"a}inen.
\newblock Gender slopes: Counterfactual fairness for computer vision models by attribute manipulation.
\newblock In \emph{FATE/MM}, pages 1--5, 2020.

\bibitem[Jung et~al.(2022)Jung, Chun, and Moon]{jung2022learning}
Sangwon Jung, Sanghyuk Chun, and Taesup Moon.
\newblock Learning fair classifiers with partially annotated group labels.
\newblock In \emph{CVPR}, pages 10348--10357, 2022.

\bibitem[Kaplun et~al.(2023)Kaplun, Gurevich, Swisa, David, Shalev-Shwartz, and Malach]{kaplun2023less}
Gal Kaplun, Andrey Gurevich, Tal Swisa, Mazor David, Shai Shalev-Shwartz, and Eran Malach.
\newblock Less is more: Selective layer finetuning with subtuning.
\newblock \emph{arXiv preprint arXiv:2302.06354}, 2023.

\bibitem[Kappiyath et~al.(2025)Kappiyath, Chaudhuri, JAISWAL, Liu, Li, Zhu, and Yin]{kappiyath2025sebra}
Adarsh Kappiyath, Abhra Chaudhuri, AJAY~KUMAR JAISWAL, Ziquan Liu, Yunpeng Li, Xiatian Zhu, and Lu Yin.
\newblock Sebra: Debiasing through self-guided bias ranking.
\newblock In \emph{ICLR}, pages 1--12, 2025.

\bibitem[Kornblith et~al.(2019)Kornblith, Shlens, and Le]{kornblith2019better}
Simon Kornblith, Jonathon Shlens, and Quoc~V Le.
\newblock Do better imagenet models transfer better?
\newblock In \emph{CVPR}, pages 2661--2671, 2019.

\bibitem[Lee et~al.(2022)Lee, Chen, Tajwar, Kumar, Yao, Liang, and Finn]{leesurgical}
Yoonho Lee, Annie~S Chen, Fahim Tajwar, Ananya Kumar, Huaxiu Yao, Percy Liang, and Chelsea Finn.
\newblock Surgical fine-tuning improves adaptation to distribution shifts.
\newblock In \emph{ICLR}, 2022.

\bibitem[Levy et~al.(2020)Levy, Carmon, Duchi, and Sidford]{levy2020large}
Daniel Levy, Yair Carmon, John~C Duchi, and Aaron Sidford.
\newblock Large-scale methods for distributionally robust optimization.
\newblock \emph{NeurIPS}, 33:\penalty0 8847--8860, 2020.

\bibitem[Liu et~al.(2021)Liu, Haghgoo, Chen, Raghunathan, Koh, Sagawa, Liang, and Finn]{liu2021just}
Evan~Z Liu, Behzad Haghgoo, Annie~S Chen, Aditi Raghunathan, Pang~Wei Koh, Shiori Sagawa, Percy Liang, and Chelsea Finn.
\newblock Just train twice: Improving group robustness without training group information.
\newblock In \emph{ICML}, pages 6781--6792, 2021.

\bibitem[Liu et~al.(2019)Liu, Simchowitz, and Hardt]{liu2019implicit}
Lydia~T Liu, Max Simchowitz, and Moritz Hardt.
\newblock The implicit fairness criterion of unconstrained learning.
\newblock In \emph{ICML}, pages 4051--4060. PMLR, 2019.

\bibitem[Liu et~al.(2015)Liu, Luo, Wang, and Tang]{liu2015faceattributes}
Ziwei Liu, Ping Luo, Xiaogang Wang, and Xiaoou Tang.
\newblock Deep learning face attributes in the wild.
\newblock In \emph{ICCV}, 2015.

\bibitem[Liu et~al.(2023)Liu, Xu, Ji, and Chan]{liu2023twins}
Ziquan Liu, Yi Xu, Xiangyang Ji, and Antoni~B Chan.
\newblock Twins: A fine-tuning framework for improved transferability of adversarial robustness and generalization.
\newblock In \emph{CVPR}, pages 16436--16446, 2023.

\bibitem[Lopez et~al.(2018)Lopez, Regier, Jordan, and Yosef]{lopez2018information}
Romain Lopez, Jeffrey Regier, Michael~I Jordan, and Nir Yosef.
\newblock Information constraints on auto-encoding variational bayes.
\newblock \emph{NeurIPS}, 31, 2018.

\bibitem[Madras et~al.(2018)Madras, Creager, Pitassi, and Zemel]{madras2018learning}
David Madras, Elliot Creager, Toniann Pitassi, and Richard Zemel.
\newblock Learning adversarially fair and transferable representations.
\newblock In \emph{ICML}, pages 3384--3393, 2018.

\bibitem[Mehrabi et~al.(2021)Mehrabi, Morstatter, Saxena, Lerman, and Galstyan]{mehrabi2021survey}
Ninareh Mehrabi, Fred Morstatter, Nripsuta Saxena, Kristina Lerman, and Aram Galstyan.
\newblock A survey on bias and fairness in machine learning.
\newblock \emph{ACM Computing Surveys}, 54\penalty0 (6):\penalty0 1--35, 2021.

\bibitem[Menon and Vondrick(2022)]{menonvisual}
Sachit Menon and Carl Vondrick.
\newblock Visual classification via description from large language models.
\newblock In \emph{ICLR}, 2022.

\bibitem[Nam et~al.(2020)Nam, Cha, Ahn, Lee, and Shin]{nam2020learning}
Junhyun Nam, Hyuntak Cha, Sungsoo Ahn, Jaeho Lee, and Jinwoo Shin.
\newblock Learning from failure: De-biasing classifier from biased classifier.
\newblock \emph{NeurIPS}, 33:\penalty0 20673--20684, 2020.

\bibitem[Oldfield et~al.(2025)Oldfield, Georgopoulos, Chrysos, Tzelepis, Panagakis, Nicolaou, Deng, and Patras]{oldfield2025multilinear}
James Oldfield, Markos Georgopoulos, Grigorios Chrysos, Christos Tzelepis, Yannis Panagakis, Mihalis Nicolaou, Jiankang Deng, and Ioannis Patras.
\newblock Multilinear mixture of experts: Scalable expert specialization through factorization.
\newblock \emph{NeurIPS}, 37:\penalty0 53022--53063, 2025.

\bibitem[Park et~al.(2021)Park, Hwang, Kim, and Byun]{park2021learning}
Sungho Park, Sunhee Hwang, Dohyung Kim, and Hyeran Byun.
\newblock Learning disentangled representation for fair facial attribute classification via fairness-aware information alignment.
\newblock In \emph{AAAI}, pages 2403--2411, 2021.

\bibitem[Park et~al.(2022)Park, Lee, Lee, Hwang, Kim, and Byun]{park2022fair}
Sungho Park, Jewook Lee, Pilhyeon Lee, Sunhee Hwang, Dohyung Kim, and Hyeran Byun.
\newblock Fair contrastive learning for facial attribute classification.
\newblock In \emph{CVPR}, pages 10389--10398, 2022.

\bibitem[Peychev et~al.(2022)Peychev, Ruoss, Balunovi{\'c}, Baader, and Vechev]{peychev2022latent}
Momchil Peychev, Anian Ruoss, Mislav Balunovi{\'c}, Maximilian Baader, and Martin Vechev.
\newblock Latent space smoothing for individually fair representations.
\newblock In \emph{ECCV}, pages 535--554, 2022.

\bibitem[Pezeshki et~al.(2024)Pezeshki, Bouchacourt, Ibrahim, Ballas, Vincent, and Lopez-Paz]{pezeshkidiscovering}
Mohammad Pezeshki, Diane Bouchacourt, Mark Ibrahim, Nicolas Ballas, Pascal Vincent, and David Lopez-Paz.
\newblock Discovering environments with xrm.
\newblock In \emph{ICML}, 2024.

\bibitem[Pleiss et~al.(2017)Pleiss, Raghavan, Wu, Kleinberg, and Weinberger]{pleiss2017fairness}
Geoff Pleiss, Manish Raghavan, Felix Wu, Jon Kleinberg, and Kilian~Q Weinberger.
\newblock On fairness and calibration.
\newblock \emph{NeurIPS}, 30, 2017.

\bibitem[Radford et~al.(2021)Radford, Kim, Hallacy, Ramesh, Goh, Agarwal, Sastry, Askell, Mishkin, Clark, et~al.]{radford2021learning}
Alec Radford, Jong~Wook Kim, Chris Hallacy, Aditya Ramesh, Gabriel Goh, Sandhini Agarwal, Girish Sastry, Amanda Askell, Pamela Mishkin, Jack Clark, et~al.
\newblock Learning transferable visual models from natural language supervision.
\newblock In \emph{ICML}, pages 8748--8763, 2021.

\bibitem[Ramaswamy et~al.(2021)Ramaswamy, Kim, and Russakovsky]{ramaswamy2021fair}
Vikram~V Ramaswamy, Sunnie~SY Kim, and Olga Russakovsky.
\newblock Fair attribute classification through latent space de-biasing.
\newblock In \emph{CVPR}, pages 9301--9310, 2021.

\bibitem[Ramesh et~al.(2022)Ramesh, Dhariwal, Nichol, Chu, and Chen]{ramesh2022hierarchical}
Aditya Ramesh, Prafulla Dhariwal, Alex Nichol, Casey Chu, and Mark Chen.
\newblock Hierarchical text-conditional image generation with clip latents.
\newblock \emph{arXiv preprint arXiv:2204.06125}, pages 1--27, 2022.

\bibitem[Ro and Choi(2021)]{ro2021autolr}
Youngmin Ro and Jin~Young Choi.
\newblock Autolr: Layer-wise pruning and auto-tuning of learning rates in fine-tuning of deep networks.
\newblock In \emph{AAAI}, pages 2486--2494, 2021.

\bibitem[Rombach et~al.(2022)Rombach, Blattmann, Lorenz, Esser, and Ommer]{rombach2022high}
Robin Rombach, Andreas Blattmann, Dominik Lorenz, Patrick Esser, and Bj{\"o}rn Ommer.
\newblock High-resolution image synthesis with latent diffusion models.
\newblock In \emph{CVPR}, pages 10684--10695, 2022.

\bibitem[Ronneberger et~al.(2015)Ronneberger, Fischer, and Brox]{ronneberger2015u}
Olaf Ronneberger, Philipp Fischer, and Thomas Brox.
\newblock U-net: Convolutional networks for biomedical image segmentation.
\newblock In \emph{MICCAI}, pages 234--241, 2015.

\bibitem[Saharia et~al.(2022{\natexlab{a}})Saharia, Chan, Saxena, Li, Whang, Denton, Ghasemipour, Gontijo~Lopes, Karagol~Ayan, Salimans, et~al.]{saharia2022}
Chitwan Saharia, William Chan, Saurabh Saxena, Lala Li, Jay Whang, Emily~L Denton, Kamyar Ghasemipour, Raphael Gontijo~Lopes, Burcu Karagol~Ayan, Tim Salimans, et~al.
\newblock Photorealistic text-to-image diffusion models with deep language understanding.
\newblock In \emph{NeurIPS}, pages 36479--36494, 2022{\natexlab{a}}.

\bibitem[Saharia et~al.(2022{\natexlab{b}})Saharia, Chan, Saxena, Li, Whang, Denton, Ghasemipour, Gontijo~Lopes, Karagol~Ayan, Salimans, et~al.]{saharia2022photorealistic}
Chitwan Saharia, William Chan, Saurabh Saxena, Lala Li, Jay Whang, Emily~L Denton, Kamyar Ghasemipour, Raphael Gontijo~Lopes, Burcu Karagol~Ayan, Tim Salimans, et~al.
\newblock Photorealistic text-to-image diffusion models with deep language understanding.
\newblock \emph{NeurIPS}, 35:\penalty0 36479--36494, 2022{\natexlab{b}}.

\bibitem[Shin et~al.(2023)Shin, Kang, and Park]{shin2023fill}
Joonghyuk Shin, Minguk Kang, and Jaesik Park.
\newblock Fill-up: Balancing long-tailed data with generative models.
\newblock \emph{arXiv preprint arXiv:2306.07200}, 2023.

\bibitem[Shrestha et~al.(2024)Shrestha, Zou, Chen, Li, Xie, and Deng]{shrestha2024fairrag}
Robik Shrestha, Yang Zou, Qiuyu Chen, Zhiheng Li, Yusheng Xie, and Siqi Deng.
\newblock Fairrag: Fair human generation via fair retrieval augmentation.
\newblock In \emph{CVPR}, pages 11996--12005, 2024.

\bibitem[Shumailov et~al.(2024)Shumailov, Shumaylov, Zhao, Papernot, Anderson, and Gal]{shumailov2024ai}
Ilia Shumailov, Zakhar Shumaylov, Yiren Zhao, Nicolas Papernot, Ross Anderson, and Yarin Gal.
\newblock Ai models collapse when trained on recursively generated data.
\newblock \emph{Nature}, 631\penalty0 (8022):\penalty0 755--759, 2024.

\bibitem[Singh et~al.(2024)Singh, Navaratnam, Holmer, Schaub-Meyer, and Roth]{singh2024synthetic}
Krishnakant Singh, Thanush Navaratnam, Jannik Holmer, Simone Schaub-Meyer, and Stefan Roth.
\newblock Is synthetic data all we need? benchmarking the robustness of models trained with synthetic images.
\newblock In \emph{CVPR}, pages 2505--2515, 2024.

\bibitem[Uelwer et~al.(2023)Uelwer, Robine, Wagner, H{\"o}ftmann, Upschulte, Konietzny, Behrendt, and Harmeling]{uelwer2023survey}
Tobias Uelwer, Jan Robine, Stefan~Sylvius Wagner, Marc H{\"o}ftmann, Eric Upschulte, Sebastian Konietzny, Maike Behrendt, and Stefan Harmeling.
\newblock A survey on self-supervised representation learning.
\newblock \emph{arXiv preprint arXiv:2308.11455}, 2023.

\bibitem[Wang and Deng(2020)]{wang2020mitigating}
Mei Wang and Weihong Deng.
\newblock Mitigating bias in face recognition using skewness-aware reinforcement learning.
\newblock In \emph{CVPR}, pages 9322--9331, 2020.

\bibitem[Wang et~al.(2024)Wang, Chung, Wu, and De~la Torre]{wang2024domain}
Yinong~Oliver Wang, Younjoon Chung, Chen~Henry Wu, and Fernando De~la Torre.
\newblock Domain gap embeddings for generative dataset augmentation.
\newblock In \emph{CVPR}, pages 28684--28694, 2024.

\bibitem[Wang et~al.(2020)Wang, Qinami, Karakozis, Genova, Nair, Hata, and Russakovsky]{wang2020towards}
Zeyu Wang, Klint Qinami, Ioannis~Christos Karakozis, Kyle Genova, Prem Nair, Kenji Hata, and Olga Russakovsky.
\newblock Towards fairness in visual recognition: Effective strategies for bias mitigation.
\newblock In \emph{CVPR}, pages 8919--8928, 2020.

\bibitem[Xue et~al.(2024)Xue, Yan, Dutt, Haider, Liu, McDonagh, and Tsaftaris]{xue2024bmft}
Yuyang Xue, Junyu Yan, Raman Dutt, Fasih Haider, Jingshuai Liu, Steven McDonagh, and Sotirios~A Tsaftaris.
\newblock Bmft: Achieving fairness via bias-based weight masking fine-tuning.
\newblock In \emph{MICCAI Workshop on Fairness of AI in Medical Imaging}, pages 98--108, 2024.

\bibitem[Ye-Bin et~al.(2023)Ye-Bin, Hyeon-Woo, Choi, Kim, Kwak, and Oh]{ye2023exploiting}
Moon Ye-Bin, Nam Hyeon-Woo, Wonseok Choi, Nayeong Kim, Suha Kwak, and Tae-Hyun Oh.
\newblock Exploiting synthetic data for data imbalance problems: Baselines from a data perspective.
\newblock \emph{arXiv preprint arXiv:2308.00994}, 2023.

\bibitem[Zhang et~al.(2023{\natexlab{a}})Zhang, Chen, Chai, Wu, Lagun, Beeler, and De~la Torre]{zhang2023iti}
Cheng Zhang, Xuanbai Chen, Siqi Chai, Chen~Henry Wu, Dmitry Lagun, Thabo Beeler, and Fernando De~la Torre.
\newblock Iti-gen: Inclusive text-to-image generation.
\newblock In \emph{ICCV}, pages 3969--3980, 2023{\natexlab{a}}.

\bibitem[Zhang et~al.(2023{\natexlab{b}})Zhang, Kuang, Chen, Liu, Wu, and Xiao]{zhang2023fairness}
Fengda Zhang, Kun Kuang, Long Chen, Yuxuan Liu, Chao Wu, and Jun Xiao.
\newblock Fairness-aware contrastive learning with partially annotated sensitive attributes.
\newblock In \emph{ICLR}, 2023{\natexlab{b}}.

\bibitem[Zhang et~al.(2024)Zhang, He, Kuang, Liu, Chen, Wu, Xiao, and Zhang]{zhang2024distributionally}
Fengda Zhang, Qianpei He, Kun Kuang, Jiashuo Liu, Long Chen, Chao Wu, Jun Xiao, and Hanwang Zhang.
\newblock Distributionally generative augmentation for fair facial attribute classification.
\newblock In \emph{CVPR}, pages 22797--22808, 2024.

\bibitem[Zhang et~al.(2017)Zhang, Song, and Qi]{zhifei2017cvpr}
Zhifei Zhang, Yang Song, and Hairong Qi.
\newblock Age progression/regression by conditional adversarial autoencoder.
\newblock In \emph{CVPR}, 2017.

\bibitem[Zhao et~al.(2024)Zhao, Dai, Li, Hu, Zhang, and Liu]{zhao2024ltgc}
Qihao Zhao, Yalun Dai, Hao Li, Wei Hu, Fan Zhang, and Jun Liu.
\newblock Ltgc: Long-tail recognition via leveraging llms-driven generated content.
\newblock In \emph{CVPR}, pages 19510--19520, 2024.

\bibitem[Zhao and Patras(2023)]{zhao2023dferclip}
Zengqun Zhao and Ioannis Patras.
\newblock Prompting visual-language models for dynamic facial expression recognition.
\newblock In \emph{BMVC}, pages 1--14, 2023.

\bibitem[Zhou et~al.(2022)Zhou, Lin, Pi, Zhang, Xu, Cui, and Zhang]{zhou2022model}
Xiao Zhou, Yong Lin, Renjie Pi, Weizhong Zhang, Renzhe Xu, Peng Cui, and Tong Zhang.
\newblock Model agnostic sample reweighting for out-of-distribution learning.
\newblock In \emph{ICML}, pages 27203--27221, 2022.

\bibitem[Zhou et~al.(2023)Zhou, Sahak, and Ba]{zhou2023using}
Yongchao Zhou, Hshmat Sahak, and Jimmy Ba.
\newblock Using synthetic data for data augmentation to improve classification accuracy.
\newblock In \emph{ICML}, 2023.

\bibitem[Zhuang et~al.(2020)Zhuang, Qi, Duan, Xi, Zhu, Zhu, Xiong, and He]{zhuang2020comprehensive}
Fuzhen Zhuang, Zhiyuan Qi, Keyu Duan, Dongbo Xi, Yongchun Zhu, Hengshu Zhu, Hui Xiong, and Qing He.
\newblock A comprehensive survey on transfer learning.
\newblock \emph{Proceedings of the IEEE}, 109\penalty0 (1):\penalty0 43--76, 2020.

\end{thebibliography}

\maketitlesupplementary

\section{Addition Details of the AIM-Fair}
\textbf{Details for Instructing LLM.} For the CelebA dataset \cite{liu2015faceattributes}, we use the following instruction as input for GPT-4, with all other attributes derived from the CelebA dataset:

\textit{Generate \{Number\} diverse text prompts for human face attributes that always include \{Target Attribute\} and \{Protected Attribute\}. Each prompt should also consider some of the following attributes: 5 o’clock shadow, arched eyebrows, attractive, bags under eyes, bald, bangs, big lips, big nose, black hair, blurry, brown hair, bushy eyebrows, chubby, double chin, eyeglasses, goatee, grey hair, heavy makeup, high cheekbones, mouth slightly open, moustache, narrow eyes, no beard, oval face, pale skin, pointy nose, receding hairline, rosy cheeks, sideburns, straight hair, wavy hair, wearing earrings, wearing a hat, wearing lipstick, wearing necklace, wearing necktie, and young. Additionally, each prompt should include a variety of head poses, such as slight tilts, turns, and different head orientations (e.g., head turned slightly left, tilted upward, facing slightly downward) to ensure diversity in the generated image angles. The prompts should be for a “Portrait face photo of a,” ensuring the image only contains the head part.}

For the UTKFace dataset \cite{zhifei2017cvpr}, given the broad age distribution, we use the following instruction for GPT-4:

\textit{Generate \{Number\} diverse text prompts for human face attributes that always include \{Target Attribute\} and \{Protected Attribute\}. Include details about facial expressions, hairstyles, and any other distinguishing features that can help in generating a realistic image. And also contain age start from 1 and end at 100. The prompts should be for a “Portrait face photo of a,” ensuring the image only contains the face part.}

\begin{figure*}[!b]
    \centering
    \includegraphics[scale=0.55]{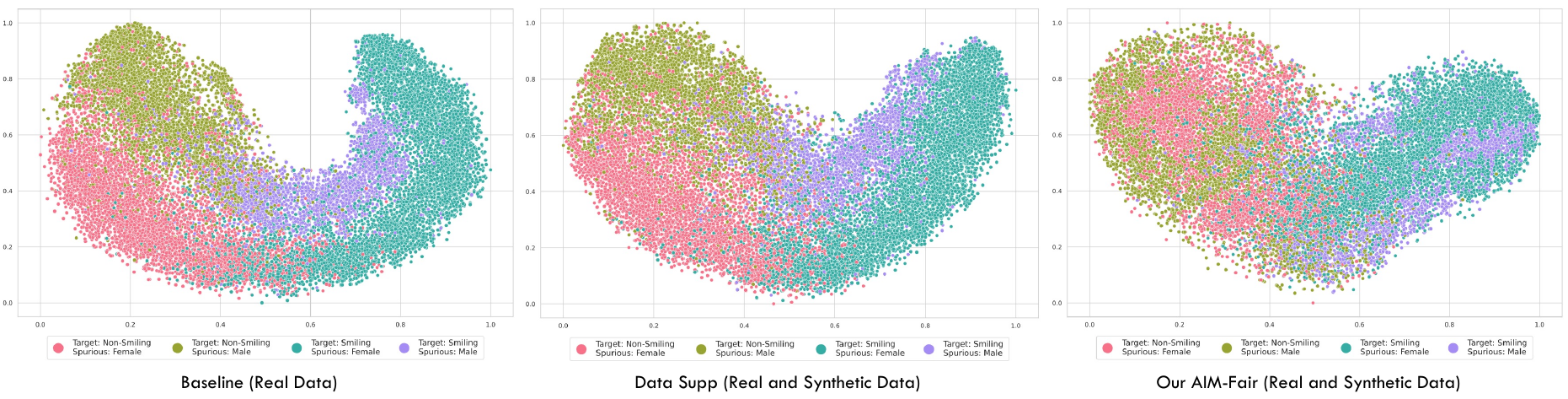}
\caption{T-SNE visualizations for the learned representations on CelebA with target attribute \textbf{\textit{Smiling}} and protected attribute \textbf{\textit{Male}}.}
\label{fig_visual_tsne}
\end{figure*}

\noindent \textbf{Algorithm for the Selective Fine-Tuning.} The algorithm of the proposed selective fine-tuning method is depicted in \cref{a1}.

\begin{algorithm}[!t]
\small
\caption{Pseudo Code for the Selective Fine-Tuning}
\label{a1}
\begin{algorithmic}

\Require Pre-trained model $f_{\theta}(x)$; Unfair real dataset $(X_{R}, Y_{R}) \in  \mathcal{D}_{R} $; 
Unfair synthetic dataset $(X_{S_1}, Y_{S_1}) \in \mathcal{D}_{S_1} $;
Fair synthetic dataset $(X_{S_2}, Y_{S_2}) \in \mathcal{D}_{S_2}  $; Top-$k$ value $k$.
\Ensure Fine-tuned model $f_{\theta^*}(x)$

\State \textbf{1. Compute Gradients:}
\State $\textbf{\textit{g}}_R \gets \nabla_{\theta} \mathcal{L}(f_{\theta}(X_R), Y_R)$
\State $\textbf{\textit{g}}_{S_1} \gets \nabla_{\theta} \mathcal{L}(f_{\theta}(X_{S1}), Y_{S_1})$
\State $\textbf{\textit{g}}_{S_2} \gets \nabla_{\theta} \mathcal{L}(f_{\theta}(X_{S2}), Y_{S_2})$

\State \textbf{2. Calculate Gradient Differences:}
\State $\Delta_1 \gets |\textbf{\textit{g}}_R - \textbf{\textit{g}}_{S_1}|$, \quad $\Delta_2 \gets |\textbf{\textit{g}}_{S_1} - \textbf{\textit{g}}_{S_2}|$

\State \textbf{3. Select Parameters to Fine-Tune:}
\State $R_1 \gets$ parameters sorted ascendingly by $\Delta_1$
\State $R_2 \gets$ parameters sorted descendingly by $\Delta_2$
\State $K \gets$ top-$k$ parameters in $R_1$ $\cap$ top-$k$ parameters in $R_2$

\State \textbf{4. Update Model Parameters:}
\For{each parameter $\theta_j$}
    \If{$\theta_i \in K$}
        \State Set $\theta_j\text{requires\_grad} \gets \text{True}$
    \Else
        \State Set $\theta_j.\text{requires\_grad} \gets \text{False}$
    \EndIf
\EndFor
\State \textbf{5. Fine-tune the Model:}
\State Train $f_{\theta}(x)$ on $(X_{S2}, Y_{S2})$ using parameters in $K$
\end{algorithmic}
\end{algorithm}

\section{Additional Experimental Results}
\textbf{Ablation Analysis on ViT Model.} We also applied our method to the ViT-32-Small model on CelebA. As shown in Tab.~\ref{results_with_vit}, our approach consistently outperforms other training methods, achieving the highest overall accuracy and the best worst-group accuracy. This demonstrates its generalization capability across different models. We also experimented by comparing our selective fine-tuning with LoRA \cite{hu2022lora} fine-tuning, and as shown in Tab. \ref{results_with_vit}, our approach consistently outperforms LoRA on both datasets. We believe that LoRA is effective for parameter-efficient adaptation but restricts the model’s ability to correct deeply embedded biases, while our method selectively fine-tunes the layers that contribute most to bias, ensuring targeted fairness improvements. Additionally, we compare our method with LTGC \cite{zhao2024ltgc} which was trained with a mix-up of real and synthetic data for addressing the long-tail problem. As shown in the results in Tab.~\ref{tab_diff_training_resnet}, while the mix-up method successfully addresses long-tail issues, it does not appear to enhance model fairness.

\noindent \textbf{Evaluations of Different Number of Synthetic Data.} We also evaluate fine-tuning with varying amounts of balanced synthetic data on UTKFace. Similar to the experiment on the CelebA, we use the real training data count as a reference and set different ratios to determine the number of synthetic data. As shown in Tab.~\ref{tab_diff_real_syn_data}, the results demonstrate the consistency with our observations on the CelebA. 

\begin{table*}[!t]
\begin{center}
\caption{Comparisons of varied training strategies with ViT-32-Small on CelebA.}
\vspace{-1em}
\label{results_with_vit}
\setlength{\tabcolsep}{2pt}
\resizebox{2\columnwidth}{!}{%
\begin{tabular}{@{}c|ccccccc|ccccccc@{}}
\toprule
\multirow{3}{*}{Methods} & \multicolumn{7}{c|}{T: Smiling; P: Male} & \multicolumn{7}{c}{T: Smiling; P: Young} \\ \cmidrule(l){2-15} 
 & \multicolumn{1}{c|}{\multirow{2}{*}{Target}} & \multicolumn{2}{c|}{Protected} & \multirow{2}{*}{ACC  (↑)} & \multirow{2}{*}{WST (↑)} & \multirow{2}{*}{EO  (↓)} & \multirow{2}{*}{STD (↓)} & \multicolumn{1}{c|}{\multirow{2}{*}{Target}} & \multicolumn{2}{c|}{Protected} & \multirow{2}{*}{ACC (↑)} & \multirow{2}{*}{WST (↑)} & \multirow{2}{*}{EO (↓)} & \multirow{2}{*}{STD (↓)} \\ \cmidrule(lr){3-4} \cmidrule(lr){10-11}
 & \multicolumn{1}{c|}{} & P=0 & \multicolumn{1}{c|}{P=1} &  &  &  &  & \multicolumn{1}{c|}{} & P=0 & \multicolumn{1}{c|}{P=1} &  &  &  &  \\ \midrule
\multirow{2}{*}{Baseline} & \multicolumn{1}{c|}{t=0} & 82.59 & \multicolumn{1}{c|}{98.31} & \multirow{2}{*}{89.00} & \multirow{2}{*}{71.12} & \multirow{2}{*}{20.18} & \multirow{2}{*}{10.91} & \multicolumn{1}{c|}{t=0} & 75.37 & \multicolumn{1}{c|}{95.41} & \multirow{2}{*}{89.32} & \multirow{2}{*}{75.37} & \multirow{2}{*}{14.61} & \multirow{2}{*}{8.17} \\
 & \multicolumn{1}{c|}{t=1} & 95.76 & \multicolumn{1}{c|}{71.12} &  &  &  &  & \multicolumn{1}{c|}{t=1} & 94.86 & \multicolumn{1}{c|}{85.69} &  &  &  &  \\ \midrule
\multirow{2}{*}{Trained on Synthetic Data} & \multicolumn{1}{c|}{t=0} & 90.10 & \multicolumn{1}{c|}{86.90} & \multirow{2}{*}{86.19} & \multirow{2}{*}{83.73} & \multirow{2}{*}{\textbf{2.32}} & \multirow{2}{*}{2.61} & \multicolumn{1}{c|}{t=0} & 82.68 & \multicolumn{1}{c|}{89.00} & \multirow{2}{*}{86.31} & \multirow{2}{*}{82.68} & \multirow{2}{*}{\textbf{3.53}} & \multirow{2}{*}{\textbf{2.27}} \\
 & \multicolumn{1}{c|}{t=1} & 83.73 & \multicolumn{1}{c|}{83.85} &  &  &  &  & \multicolumn{1}{c|}{t=1} & 85.01 & \multicolumn{1}{c|}{85.16} &  &  &  &  \\ \midrule
\multirow{2}{*}{Data Supplementation} & \multicolumn{1}{c|}{t=0} & 81.54 & \multicolumn{1}{c|}{97.89} & \multirow{2}{*}{88.97} & \multirow{2}{*}{72.89} & \multirow{2}{*}{19.72} & \multirow{2}{*}{10.35} & \multicolumn{1}{c|}{t=0} & 76.43 & \multicolumn{1}{c|}{95.39} & \multirow{2}{*}{89.47} & \multirow{2}{*}{76.43} & \multirow{2}{*}{13.75} & \multirow{2}{*}{7.65} \\
 & \multicolumn{1}{c|}{t=1} & 95.99 & \multicolumn{1}{c|}{72.89} &  &  &  &  & \multicolumn{1}{c|}{t=1} & 94.43 & \multicolumn{1}{c|}{85.90} &  &  &  &  \\ \midrule
\multirow{2}{*}{Fully Fine-Tuning} & \multicolumn{1}{c|}{t=0} & 89.68 & \multicolumn{1}{c|}{91.68} & \multirow{2}{*}{89.37} & \multirow{2}{*}{83.95} & \multirow{2}{*}{4.17} & \multirow{2}{*}{2.92} & \multicolumn{1}{c|}{t=0} & 83.83 & \multicolumn{1}{c|}{92.24} & \multirow{2}{*}{88.80} & \multirow{2}{*}{{\ul 83.83}} & \multirow{2}{*}{5.28} & \multirow{2}{*}{3.06} \\
 & \multicolumn{1}{c|}{t=1} & 90.01 & \multicolumn{1}{c|}{83.95} &  &  &  &  & \multicolumn{1}{c|}{t=1} & 88.86 & \multicolumn{1}{c|}{86.85} &  &  &  &  \\ \midrule
\multirow{2}{*}{LoRA \cite{hu2022lora}} & \multicolumn{1}{c|}{t=0} & 91.20 & \multicolumn{1}{c|}{91.93} & \multirow{2}{*}{{\ul 90.79}} & \multirow{2}{*}{{\ul 86.20}} & \multirow{2}{*}{{\ul 3.14}} & \multirow{2}{*}{{\ul 2.37}} & \multicolumn{1}{c|}{t=0} & 83.38 & \multicolumn{1}{c|}{91.15} & \multirow{2}{*}{{\ul 89.81}} & \multirow{2}{*}{83.38} & \multirow{2}{*}{5.29} & \multirow{2}{*}{3.47} \\
 & \multicolumn{1}{c|}{t=1} & 91.76 & \multicolumn{1}{c|}{86.20} &  &  &  &  & \multicolumn{1}{c|}{t=1} & 92.41 & \multicolumn{1}{c|}{89.61} &  &  &  &  \\ \midrule
\multirow{2}{*}{LTGC \cite{zhao2024ltgc}} & \multicolumn{1}{c|}{t=0} & 78.17 & \multicolumn{1}{c|}{95.20} & \multirow{2}{*}{88.75} & \multirow{2}{*}{78.17} & \multirow{2}{*}{17.20} & \multirow{2}{*}{8.63} & \multicolumn{1}{c|}{t=0} & 80.20 & \multicolumn{1}{c|}{96.64} & \multirow{2}{*}{89.18} & \multirow{2}{*}{80.20} & \multirow{2}{*}{13.20} & \multirow{2}{*}{6.80} \\
 & \multicolumn{1}{c|}{t=1} & 96.82 & \multicolumn{1}{c|}{79.44} &  &  &  &  & \multicolumn{1}{c|}{t=1} & 93.09 & \multicolumn{1}{c|}{83.13} &  &  &  &  \\ \midrule
\multirow{2}{*}{Selective Fine-Tuning} & \multicolumn{1}{c|}{t=0} & 89.90 & \multicolumn{1}{c|}{91.20} & \multirow{2}{*}{\textbf{90.89}} & \multirow{2}{*}{\textbf{87.79}} & \multirow{2}{*}{3.27} & \multirow{2}{*}{\textbf{1.85}} & \multicolumn{1}{c|}{t=0} & 84.96 & \multicolumn{1}{c|}{92.80} & \multirow{2}{*}{\textbf{90.30}} & \multirow{2}{*}{\textbf{84.96}} & \multirow{2}{*}{{\ul 5.12}} & \multirow{2}{*}{{\ul 2.98}} \\
 & \multicolumn{1}{c|}{t=1} & 92.84 & \multicolumn{1}{c|}{87.79} &  &  &  &  & \multicolumn{1}{c|}{t=1} & 91.45 & \multicolumn{1}{c|}{89.05} &  &  &  &  \\ \bottomrule
\end{tabular}
}
\vspace{-1.5em}
\end{center}
\end{table*}

\begin{table}[!t]
\begin{center}
\caption{Comparisons of varied training strategies with ResNet-18 on CelebA (T=Smiling, P=Male).}
\vspace{-1em}
\label{tab_diff_training_resnet}
\setlength{\tabcolsep}{2pt}
\resizebox{\columnwidth}{!}{%
\begin{tabular}{@{}c|c|cc|cccc@{}}
\toprule
\multirow{2}{*}{Methods} & \multirow{2}{*}{Target} & \multicolumn{2}{c|}{Protected} & \multirow{2}{*}{ACC (↑)} & \multirow{2}{*}{WST(↑)} & \multirow{2}{*}{EO (↓)} & \multirow{2}{*}{STD (↓)} \\ \cmidrule(lr){3-4}
 &  & P=0 & P=1 &  &  &  &  \\ \midrule
\multirow{2}{*}{Balseline} & t=0 & 83.57 & 98.50 & \multirow{2}{*}{\textbf{89.23}} & \multirow{2}{*}{71.52} & \multirow{2}{*}{23.84} & \multirow{2}{*}{10.65} \\
 & t=1 & 95.36 & 71.52 &  &  &  &  \\ \midrule
\multirow{2}{*}{LTGC \cite{zhao2024ltgc}} & t=0 & 79.48 & 97.93 & \multirow{2}{*}{89.05} & \multirow{2}{*}{75.42} & \multirow{2}{*}{21.27} & \multirow{2}{*}{10.03} \\
 & t=1 & 96.65 & 75.42 &  &  &  &  \\ \midrule
\multirow{2}{*}{\begin{tabular}[c]{@{}c@{}}\textbf{AIM-Fair} \\ \textbf{(Ours)}\end{tabular}} & t=0 & 88.16 & 91.40 & \multirow{2}{*}{89.02} & \multirow{2}{*}{\textbf{84.20}} & \multirow{2}{*}{\textbf{6.07}} & \multirow{2}{*}{\textbf{2.74}} \\
 & t=1 & 90.25 & 84.20 &  &  &  &  \\ \bottomrule
\end{tabular}
}
\end{center}
\vspace{-1.5em}
\end{table}

\begin{table}[!t]
\begin{center}
\caption{Classification results on UTKFace with different numbers of synthetic data.}
\vspace{-1em}
\label{tab_diff_real_syn_data}
\setlength{\tabcolsep}{2pt}
\resizebox{\columnwidth}{!}{%
\begin{tabular}{@{}cl|c|cc|cccc@{}}
\toprule
\multicolumn{2}{c|}{\multirow{2}{*}{\begin{tabular}[c]{@{}c@{}}Ratio To\\ Real Data\end{tabular}}} & \multirow{2}{*}{Target} & \multicolumn{2}{c|}{Protected} & \multirow{2}{*}{ACC (↑)} & \multirow{2}{*}{WST (↑)} & \multirow{2}{*}{EO (↓)} & \multirow{2}{*}{STD (↓)} \\ \cmidrule(lr){4-5}
\multicolumn{2}{c|}{} &  & P=0 & P=1 &  &  &  &  \\ \midrule
\multicolumn{2}{c|}{\multirow{2}{*}{Baseline}} & t=0 & 79.58 & 96.16 & \multirow{2}{*}{88.86} & \multirow{2}{*}{79.58} & \multirow{2}{*}{16.59} & \multirow{2}{*}{7.26} \\
\multicolumn{2}{c|}{} & t=1 & 95.75 & 83.96 &  &  &  &  \\ \midrule
\multicolumn{2}{c|}{\multirow{2}{*}{0.5}} & t=0 & 82.61 & 89.35 & \multirow{2}{*}{{\ul 88.91}} & \multirow{2}{*}{82.61} & \multirow{2}{*}{6.74} & \multirow{2}{*}{3.78} \\
\multicolumn{2}{c|}{} & t=1 & 92.06 & 91.61 &  &  &  &  \\
\multicolumn{2}{c|}{\multirow{2}{*}{1.0}} & t=0 & 84.26 & 89.08 & \multirow{2}{*}{88.30} & \multirow{2}{*}{\textbf{84.26}} & \multirow{2}{*}{\textbf{4.81}} & \multirow{2}{*}{\textbf{2.41}} \\
\multicolumn{2}{c|}{} & t=1 & 90.62 & 89.25 &  &  &  &  \\
\multicolumn{2}{c|}{\multirow{2}{*}{1.5}} & t=0 & 83.70 & 89.15 & \multirow{2}{*}{88.23} & \multirow{2}{*}{{\ul 83.70}} & \multirow{2}{*}{{\ul 5.45}} & \multirow{2}{*}{{\ul 2.65}} \\
\multicolumn{2}{c|}{} & t=1 & 89.84 & 90.25 &  &  &  &  \\
\multicolumn{2}{c|}{\multirow{2}{*}{2.0}} & t=0 & 83.04 & 89.76 & \multirow{2}{*}{\textbf{88.98}} & \multirow{2}{*}{83.04} & \multirow{2}{*}{6.73} & \multirow{2}{*}{3.51} \\
\multicolumn{2}{c|}{} & t=1 & 91.71 & 91.41 &  &  &  &  \\ \bottomrule
\end{tabular}
}
\end{center}
\vspace{-1.5em}
\end{table}

\begin{figure*}[!t]
    \centering
    \begin{subfigure}{\textwidth}
    \centering
    \includegraphics[scale=0.95]{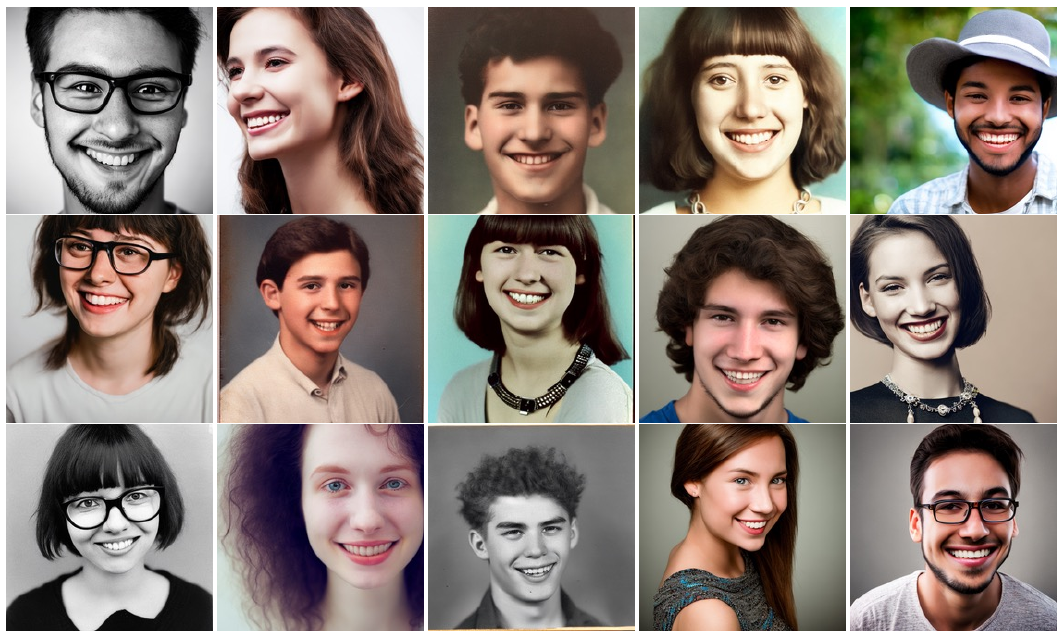}
    \caption{Generated contextual images for CelebA with target attribute \textbf{\textit{Smiling}} and protected attribute \textbf{\textit{Young}}.}
    \label{fig_celeba}
     \end{subfigure}
     \begin{subfigure}{\textwidth}
     \centering
    \includegraphics[scale=0.95]{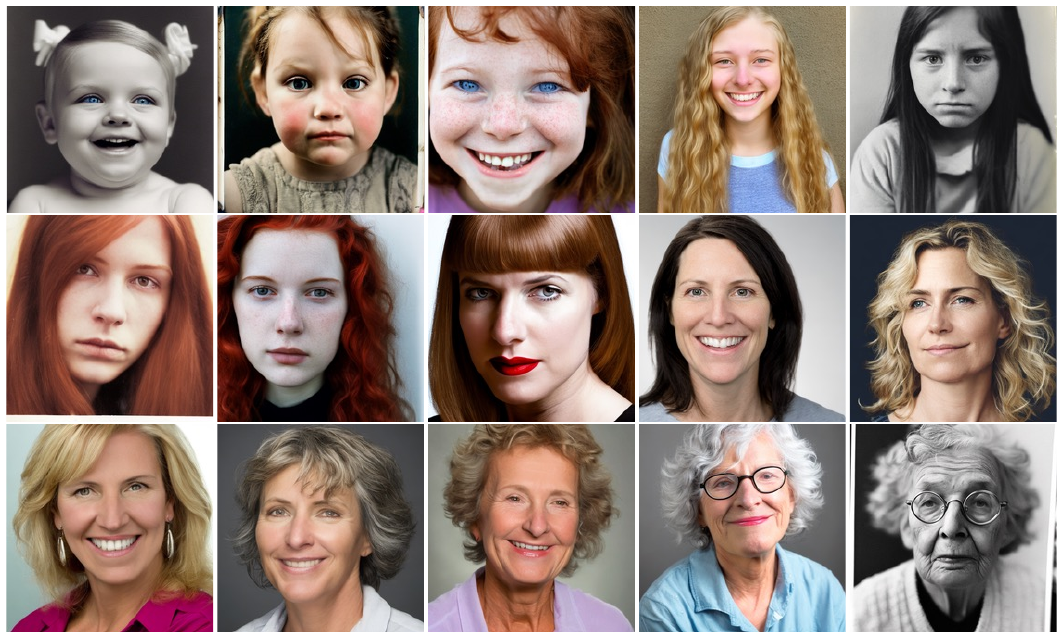}
    \caption{Generated contextual images for UTKFace with target attribute \textbf{\textit{Female}} and protected attribute \textbf{\textit{White}}.}
    \label{fig_utkface}
    \end{subfigure}
\caption{Generated contextual images.}
\label{fig_visual_images}
\end{figure*}
\begin{figure}[!t]
    \centering
    \includegraphics[width=0.45\textwidth]{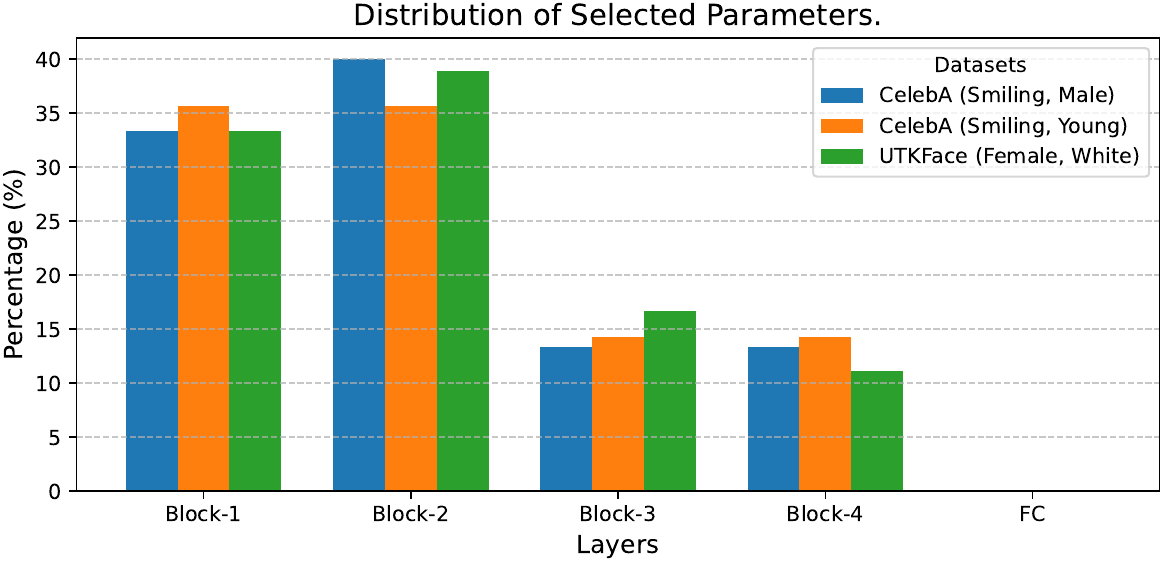}
\vspace{-1em}
\caption{Distribution of selected parameters.}
\label{mask_distribution}
\end{figure}
\section{Visualizations}
\textbf{Latent Visual Feature Distributions.} To further illustrate how our method works, we provide visualizations of the learned representations using t-SNE in \cref{fig_visual_tsne}. We divide the CelebA test set into four groups based on target and spurious attributes. We observe that the baseline ERM, trained either solely on biased real data or on a mix of real and synthetic data, learns spurious correlations, resulting in representations that can be separated by the spurious attributes. In contrast, the representations learned by our AIM-Fair method contain less information on spurious correlations, thereby contributing to fairer classification.

\noindent \textbf{More Generated Contextual Images.} We provide additional generated contextual images with target attribute \textit{Smiling} and protected attribute \textit{Young} and target attribute \textit{Female} and protected attribute \textit{White} in \cref{fig_visual_images}. \cref{fig_celeba} demonstrates that the generated images of smiling young individuals include diverse genders, hairstyles, hair colours, accessories, and head poses. \cref{fig_utkface} shows that the generated images of white females include a range of ages, facial expressions, hairstyles, hair colours, accessories, and head poses. This diversity in unmentioned attributes of the plain prompt helps mitigate bias related to these attributes.

\noindent \textbf{Generated Mask Distribution.} We analysed the distribution of the selected fine-tuned parameters across different layers. Our empirical results in Fig.~\ref{mask_distribution} show that most of the selected parameters are located in early layers – this is consistent across different datasets.

\begin{table}[!t]
\begin{center}
\caption{Classification results on CelebA dataset (T=Male, P=Young) under different training strategies.}
\vspace{-1em}
\label{tab_on_male_young}
\setlength{\tabcolsep}{6pt}
\resizebox{\columnwidth}{!}{%
\begin{tabular}{@{}c|c|cc|cccc@{}}
\toprule
\multirow{2}{*}{Methods} & \multirow{2}{*}{Target} & \multicolumn{2}{c|}{Protected} & \multirow{2}{*}{\begin{tabular}[c]{@{}c@{}}ACC\\ (↑)\end{tabular}} & \multirow{2}{*}{\begin{tabular}[c]{@{}c@{}}WST\\ (↑)\end{tabular}} & \multirow{2}{*}{\begin{tabular}[c]{@{}c@{}}EO\\ (↓)\end{tabular}} & \multirow{2}{*}{\begin{tabular}[c]{@{}c@{}}STD\\ (↓)\end{tabular}} \\ \cmidrule(lr){3-4}
 &  & P=0 & P=1 &  &  &  &  \\ \midrule
\multirow{2}{*}{Baseline} & t=0 & 93.50 & 98.80 & \multirow{2}{*}{\textbf{97.16}} & \multirow{2}{*}{93.50} & \multirow{2}{*}{5.37} & \multirow{2}{*}{2.58} \\
 & t=1 & 98.98 & 93.99 &  &  &  &  \\ \midrule
\multirow{2}{*}{\begin{tabular}[c]{@{}c@{}}Fully\\ Fine-Tuning\end{tabular}} & t=0 & 95.19 & 97.83 & \multirow{2}{*}{96.47} & \multirow{2}{*}{93.79} & \multirow{2}{*}{3.24} & \multirow{2}{*}{1.55} \\
 & t=1 & 96.88 & 93.79 &  &  &  &  \\ \midrule
\multirow{2}{*}{\begin{tabular}[c]{@{}c@{}}AIM-Fair\\ (Ours)\end{tabular}} & t=0 & 94.41 & 97.34 & \multirow{2}{*}{96.00} & \multirow{2}{*}{\textbf{93.86}} & \multirow{2}{*}{\textbf{3.00}} & \multirow{2}{*}{\textbf{1.35}} \\
 & t=1 & 95.83 & 93.86 &  &  &  &  \\ \bottomrule
\end{tabular}
}
\vspace{-2em}
\end{center}
\end{table}

\section{Failures and Limitations}
Our method still faces challenges in enhancing model fairness while retaining utility. As shown in \cref{tab_on_male_young}, the baseline results indicate that the model exhibits only a small bias on the protected attribute. In this setting, our method improves model fairness, but it sacrifices accuracy by 1.16\%. Additionally, compared to the fully fine-tuning method, our approach achieves better fairness but still results in lower accuracy. We argue that when the model achieves very high accuracy across all groups, it becomes challenging for our method to distinguish which parameters are sensitive to domain shift and which are sensitive to group shift. To address this limitation, further effort should be invested either in improving the synthetic data generation process or in designing strategies for faithful synthetic data selection.

\end{document}